\documentclass[10pt,journal,cspaper,compsoc]{IEEEtran}

\ifCLASSOPTIONcompsoc
  \usepackage[nocompress]{cite}
\else
  \usepackage{cite}
\fi

\ifCLASSINFOpdf
   \usepackage[pdftex]{graphicx}
   \graphicspath{{../pdf/}{../jpeg/}}
   \DeclareGraphicsExtensions{.pdf,.jpeg,.png,.jpg}
\else
   \usepackage[dvips]{graphicx}
   \graphicspath{{../eps/}}
   \DeclareGraphicsExtensions{.eps}
\fi

\usepackage{amsmath}
\usepackage{amsfonts}

\ifCLASSOPTIONcompsoc
  \usepackage[caption=false,font=footnotesize,labelfont=sf,textfont=sf]{subfig}
\else
  \usepackage[caption=false,font=footnotesize]{subfig}
\fi

\usepackage{stfloats}

\def\ie{{\em i.e.}}
\def\eg{{\em e.g.}}
\def\etal{{\em et al.}}

\newcommand{\figref}[1]{Fig. \ref{#1}}
\newcommand{\tabref}[1]{Tab. \ref{#1}}

\newcommand{\secref}[1]{Section \ref{#1}}

\newcommand{\mc}[1]{\mathcal{#1}}

\newcommand{\bs}[1]{\boldsymbol{\texttt{#1}}}

\usepackage{url}
\usepackage{ragged2e}
\usepackage{booktabs}
\usepackage{color}
\usepackage{multirow}
\usepackage{bm}
\usepackage{bbm}
\usepackage[misc]{ifsym}
\usepackage{graphicx}
\usepackage{amsmath}
\usepackage{amsfonts}
\usepackage{enumerate}

\usepackage{balance}
\usepackage[colorlinks,linkcolor=blue,anchorcolor=blue,citecolor=blue]{hyperref}

\usepackage{amssymb}

\usepackage[noend]{algpseudocode}
\usepackage{algorithmicx,algorithm}

\graphicspath{{./figures/}}

\begin{document}

\title{
Parsing Objects at a Finer Granularity: A Survey
}

\author{Yifan~Zhao,~Jia~Li,~\IEEEmembership{Senior Member,~IEEE},
and~Yonghong~Tian,~\IEEEmembership{Fellow,~IEEE}
\IEEEcompsocitemizethanks{
\IEEEcompsocthanksitem Y. Zhao and Y. Tian are with the School of Computer Science, Peking University, Beijing, 100871, China
\IEEEcompsocthanksitem J. Li is with the State Key Laboratory of Virtual Reality Technology and Systems, School of Computer Science and Engineering, Beihang University, Beijing, 100191, China.
\IEEEcompsocthanksitem Y. Tian is also with School of Electronic and Computer Engineering, Peking University Shenzhen Graduate School, Peking University, Shenzhen, 518055, China.

\IEEEcompsocthanksitem J. Li is the corresponding author (E-mail: jiali@buaa.edu.cn).
}}

\markboth{}%
{Shell \MakeLowercase{\textit{et al.}}:}

\IEEEtitleabstractindextext{%
\begin{abstract}
\justifying Fine-grained visual parsing, including fine-grained part segmentation and fine-grained object recognition, has attracted considerable critical attention due to its importance in many real-world applications,~\eg, agriculture, remote sensing, and space technologies. Predominant research efforts tackle these fine-grained sub-tasks following different paradigms, while the inherent relations between these tasks are neglected. Moreover, given most of the research remains fragmented, we conduct an in-depth study of the advanced work from a new perspective of learning the part relationship. In this perspective, we first consolidate recent research and benchmark syntheses with new taxonomies. Based on this consolidation, we revisit the universal challenges in fine-grained part segmentation and recognition tasks and propose new solutions by part relationship learning for these important challenges. Furthermore, we conclude several promising lines of research in fine-grained visual parsing for future research.
\end{abstract}

}

\maketitle

\IEEEdisplaynontitleabstractindextext

\IEEEpeerreviewmaketitle

\IEEEraisesectionheading{\section{Introduction}\label{sec:intro}}

\begin{figure*}[!t]
	\centering
	\includegraphics[width=0.95\linewidth]{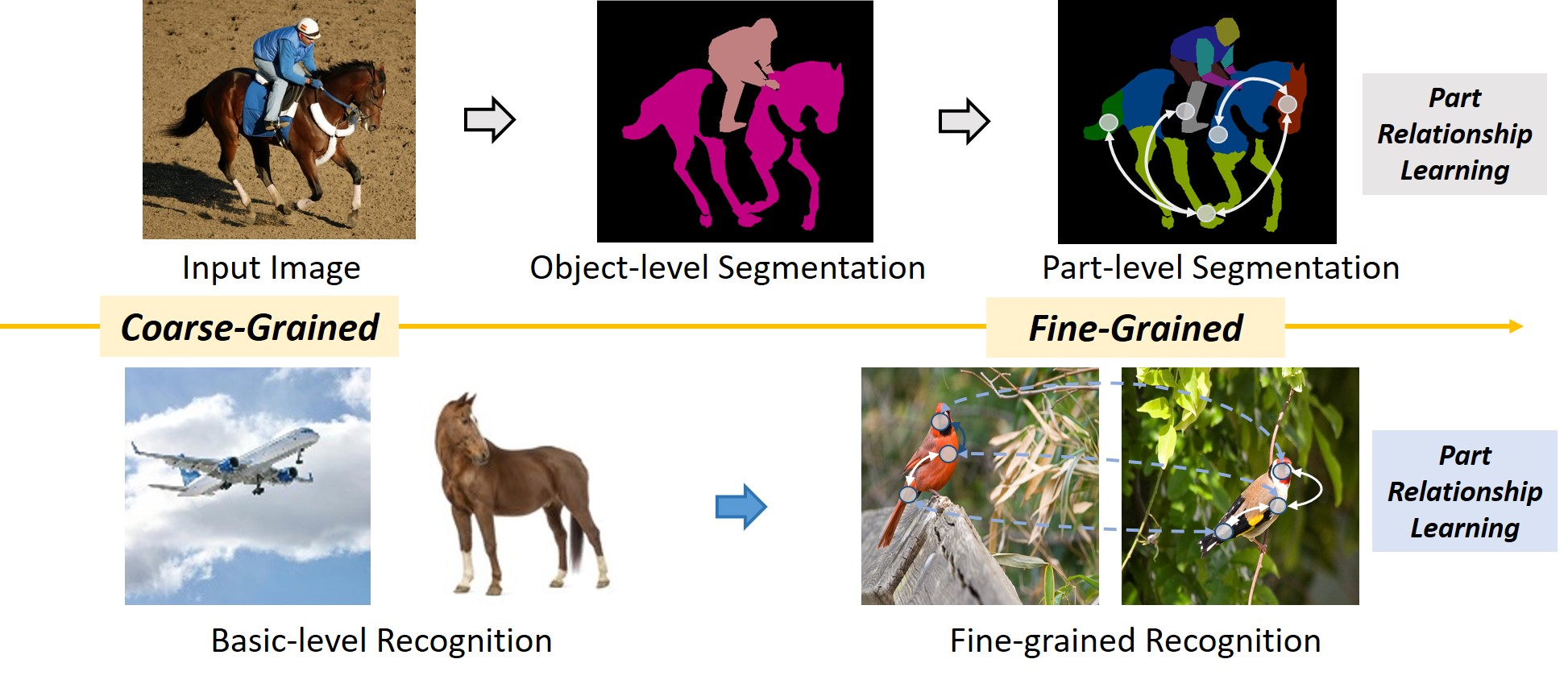}
	\caption{Comparisons of coarse-grained learning and fine-grained learning. Representative fine-grained learning tasks,~\ie, semantic part segmentation and fine-grained recognition, rely on the part relationship learning to build robust local features, while the coarse-grained tasks can be achieved by image-level global features.
	}
	\label{fig:motivation}
\end{figure*}

\IEEEPARstart{F}{ine-grained} visual parsing of image objects is a basic and crucial task in the computer vision community, which is fundamentally difficult, owing that there are usually subtle visual cues for distinguishing different objects or part regions. Recent advances in deep learning have significantly boosted the image understanding abilities of machine systems,~\eg, its performance on the large-scale ImageNet dataset~\cite{deng2009imagenet} surpasses the human-level recognition, but it is still a great challenge facing the fine-grained visual tasks. In particular, we consider two representative fine-grained visual parsing tasks in this paper,~\ie, semantic part segmentation and fine-grained object recognition.

In contrast with coarse-grained object segmentation and base-level classification, fine-grained parsing is meant to segment or distinguish visually similar objects that belong to different fine-grained concepts, for example, decomposing objects into parts and dividing the base category into subcategories.
A tremendous amount of research efforts~\cite{liang2016semantic1,ruan2019devil,zhao2021ordinal,xia2017joint,
zhang2014part,huang2016part,he2017weakly,wei2018mask,huang2020interpretable} has been proposed to solve this important problem, which can also be applied for downstream applications~\cite{bickel2020impacts,sun2022fair1m,pakhomov2019deep}. Conventional machine learning techniques build explicit structures for parsing and understanding these fine-grained objects,~\eg, graph and tree structures for part segmentation~\cite{zhu2008unsupervised,hsieh2010segmentation,wang2011learning,car2014parsing}, and part learning in fine-grained recognition~\cite{zhang2012pose,branson2011strong}. In the era of deep learning, fine-grained segmentation and recognition approaches follow different paradigms, which achieve huge success compared to conventional models. Although there are more than 100 research papers each year to investigate this important problem, these papers seem to be disorganized, owing to various sundry research focuses including new task settings, benchmarking, and learning strategies. In particular, there are few survey papers that summarize the recent advances in fine-grained part segmentation. Thus the relationships among different fine-grained sub-tasks are still under-explored, and these sub-tasks are developed independently by regarding them as less-relevant tasks.

In this paper, we make a comprehensive study of advances in fine-grained visual parsing tasks in the last decade. Besides analyzing recent deep learning works, we seek to explain the differences between non-deep learning and deep models, since these works often share similar intuitions and observations, and some of the previous studies could inspire further research. For consolidating these recent advances, we propose a new taxonomy for fine-grained part segmentation and recognition tasks, and also provide a collection of predominant benchmark datasets following our taxonomy. Besides these improvements compared to other survey papers~\cite{zhao2017survey,wei2021fine}, in this paper, we start from the novel view of~\textit{part relationship learning} and regard it as the correspondences of different fine-grained sub-tasks. In this view, we revisit both the individual and universal challenges of part segmentation and fine-grained recognition and make an attempted solution using the guidance of~\textit{part relationship learning}. In addition to these insights, we finish by discussing the future directions of fine-grained visual parsing tasks.

To summarize, the main contributions of this survey are as follows:
\begin{enumerate}
\item  We present a comprehensive survey of fine-grained visual parsing tasks by collecting recent advances of two representative tasks,~\ie, semantic part segmentation and fine-grained recognition.

\item We revisit these fine-grained visual tasks from a novel perspective of part relationship learning, by revealing the connections of these fine-grained tasks and providing a promising solution to tackle challenges in fine-grained tasks.

\item We consolidate recent fine-grained research by reorganizing these works with new taxonomy, providing a collection of prevailing benchmark datasets, and make comprehensive discussions to inspire future works.

\item We provide promising future directions of fine-grained visual parsing tasks to inform further studies.
\end{enumerate}

\begin{figure*}[!t]
	\centering
	\includegraphics[width=\linewidth]{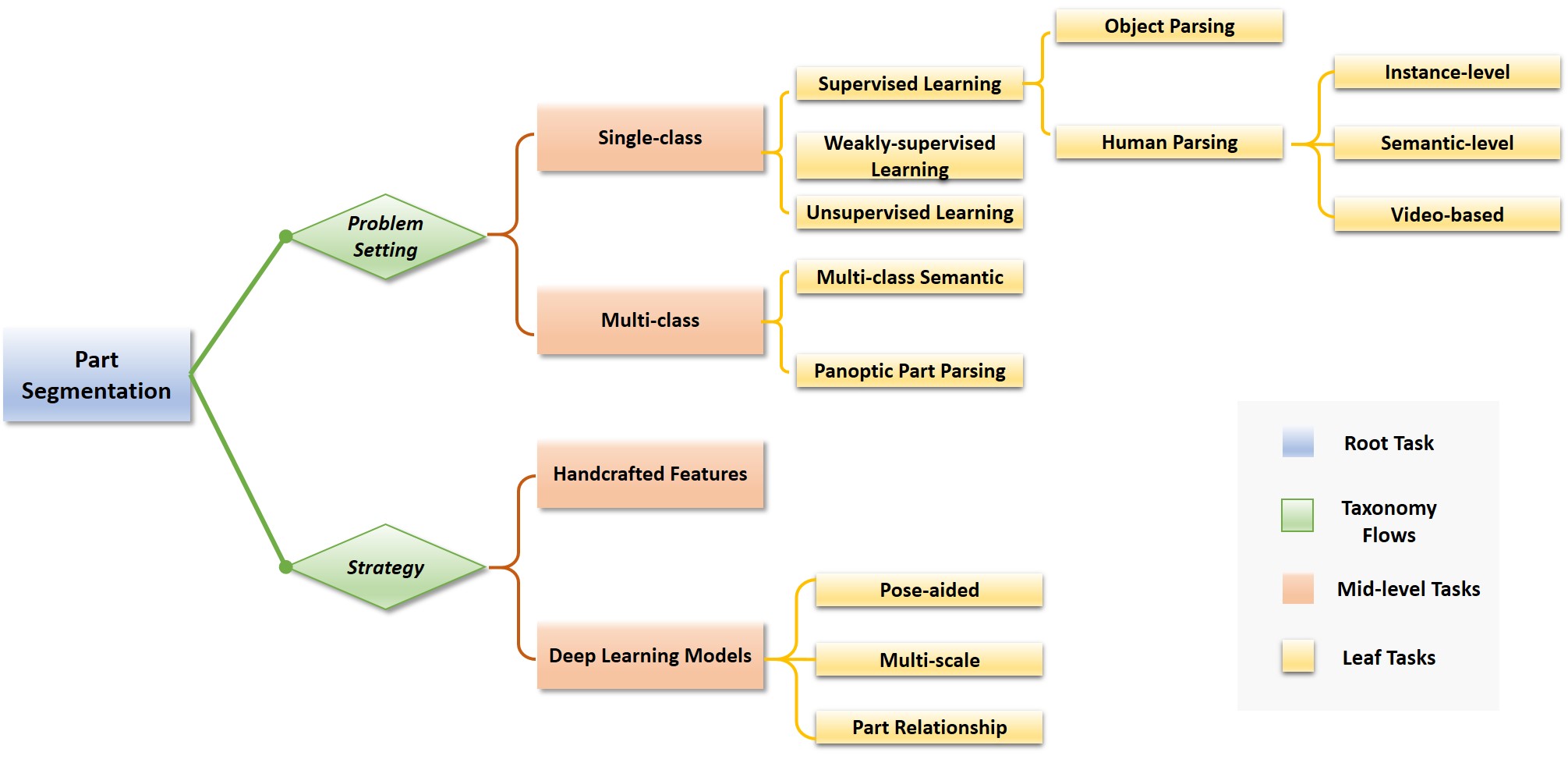}
	\caption{The landscape of semantic part segmentation tasks in our taxonomy. We summarize the recent advances from two different aspects: problem setting and learning strategy.
	}
	\label{fig:tax-part}
\end{figure*}

The remainder of this paper is organized as follows:~\secref{sec:partseg} provides new taxonomies, benchmark settings, and recent research on the problem of fine-grained part segmentation.~\secref{sec:fgvc} consolidates benchmarks, challenges, and advanced research on fine-grained object recognition. In~\secref{sec:partrelation}, we delve into the connections of different fine-grained visual tasks from the perspective of part relationship learning and provide new solutions to improve existing challenges. We then highlight the future directions in~\secref{sec:direction} and then conclude this paper in~\secref{sec:conclusion}.

\section{Fine-grained Object Segmentation: A Part-level Perspective }\label{sec:partseg}
\subsection{Taxonomy}\label{sec:tax-part}
In this section, we construct a new taxonomy for the semantic part segmentation task and revisit the fine-grained object segmentation from the part-level perspective. As in~\figref{fig:tax-part}, we conclude and re-organize the methods to solve semantic part segmentation tasks from two different views,~\ie, the problem setting and learning strategies.

1) \textbf{Problem setting.} Considering target objects to segment, we categorize these methods into two lines,~\ie, single-class and multi-class parsing. The single-class part segmentation methods only tackle one specific category while objects of other categories should be taken as backgrounds. The multi-class setting aims to segment multiple classes that appear in the visual stream simultaneously. Regarding data collection in segmentation tasks, we further divide them into strongly-supervised, weakly-supervised and unsupervised learning. In similar ways, we consider the instance-level semantic-level, and video-based parsing problems. Given these terminologies defined here, some sub-areas show linguistic crosses with other ones,~\eg, unsupervised learning with multi-class part parsing. However, as these sub-areas have not yet been explored, here we discuss the main branches that attract major research attention in~\secref{sec:task-part}.

2) \textbf{Strategy.} Learning to segment object parts has attracted a wide variety of research attention. In previous decades, several successful hand-crafted models have achieved success in segmenting objects with clear foreground representations,~\ie, salient objects. We will briefly introduce several pioneer works in the following section. Besides these hand-crafted models, deep learning techniques have substantially improved the accuracy of segmentation models. We thus roughly group these techniques into three lines,~\ie, pose-aided, multi-scale techniques, and using part relationships. Note that similar ideas could also be proposed in the non-deep learning methods. We will elaborate on their relations and differences in~\secref{sec:strategy-part}.

\begin{table*}[t]
\caption{Summarization and comparisons of 12 widely-used part segmentation benchmark datasets.  Note that PPS~\cite{de2021part} re-organizes two new datasets based on existing data annotations.  }\label{tab:part-dataset}
\centering
\renewcommand{\arraystretch}{1.4}
\resizebox{\linewidth}{!}{
\begin{tabular}{l|c|c|c|c|c|c}
\toprule
Dataset & Pub. &Year & Task & Image Num. & Category & Description\\
\midrule
Fashionista~\cite{yamaguchi2012parsing}&CVPR&2012& Human Parsing&685& 56 & Human
clothes parsing\\
PASCAL-Part~\cite{chen2014detect}&CVPR&2014&Detection \& Segmentation&10,103& NA &First large-scale part segmentation dataset\\
Horse-Cow dataset~\cite{wang2015semantic}&CVPR&2015& Single-class Part Segmentation&521& 5 &Quadruped animal parsing, reorganized from~\cite{chen2014detect}\\
ATR~\cite{liang2015human}&ICCV&2015& Human Parsing&17,700& 18 &  Human clothes parsing\\
PASCAL-Person-Part~\cite{chen2016attention}&CVPR&2016& Human Parsing&3,533& 7 & Human body parsing, reorganized from~\cite{chen2014detect}\\
MHP~\cite{li2017multiple}&arXiv&2017& Human Parsing&4,980& 18 & Multiple human clothes parsing\\
LIP~\cite{liang2018look}&T-PAMI&2018& Human Parsing&50,462& 20 & Clothes parsing with human Poses\\
VIP~\cite{zhou2018adaptive}&ACM MM&2018& Video-based Human Parsing&404 videos& 19 & Video-based human clothes parsing\\
CIHP~\cite{gong2018instance}&ECCV&2018& Instance-level Human Parsing&38,280& 20 & Human clothes parsing\\
PASCAL-Part-58~\cite{zhao2019multi}&CVPR&2019&Multi-class Part Segmentation&10,103& 58 &First multi-class dataset, reorganized from~\cite{chen2014detect}\\
PPS~\cite{de2021part}&CVPR&2021&Part-aware Panoptic Segmentation&10,103/3,475& 194/23 &Derived from VOC-2010/Cityscape dataset\\
UDA-Part~\cite{liu2022learning}&CVPR&2022&Single-class Part Segmentation&200&5&Unsupervised domain adaptation from synthetic vehicles\\
ADE-20K-Part~\cite{michieli2022edge}&IJCV&2022&Multi-class Part Segmentation&10,103& 544 &Large scale multi-class dataset, reorganized from~\cite{zhou2017scene}\\
\bottomrule
\end{tabular}
}
\end{table*}

\subsection{Task Settings in Part Segmentation}\label{sec:task-part}
Following the taxonomy in~\secref{sec:tax-part}, we first summarize these datasets according to the task settings and annotation labels. We then elaborate on the detailed task settings and popular methods to solve these problems.
According to the segmentation targets, here we summarize the popular datasets for semantic part segmentation tasks, which span the publications, image numbers, segmentation categories, and detailed descriptions.

\subsubsection{Single-class Part Segmentation}
\textbf{Human parsing.} As in~\tabref{tab:part-dataset}, earlier works first tend to solve the specific categories of part segmentation,~\ie, human parsing\footnote{Also noted as human part segmentation in some works.}. Representative datasets including Fashionista~\cite{yamaguchi2012parsing} focuses on the human clothes parsing, which segments human objects into typical classes including~\textit{shorts},~\textit{shoes},~\textit{boots} and~\textit{sweaters}. However, this dataset contains over 56 categories with a limited number of 685 images, which is not applicable to large machine-learning systems. With the development of deep learning techniques, large datasets are proposed to train and benchmark these deep models,~\eg, ATR~\cite{liang2015human} and LIP~\cite{liang2018look}, which consist of over 50,000 images of 20 categories for training and testing. These large benchmarks, as well as the accompanying baseline, have achieved great success in parsing humans into dressing clothes. Nevertheless, decomposing human objects with different clothing parts would lead to semantic inconsistencies on certain occasions. Hence the other line of works proposes to segment human bodies into semantic parts following the morphological rules, which share the same definitions with human poses. For example, Chen~\etal~\cite{chen2016attention} propose to organize the PASCAL-Person-Part dataset to segment human bodies into 7 semantic parts, including lower/upper-arms, torsos, lower/upper-legs, heads, and backgrounds. Leading by this trend, dozens of works~\cite{liang2016semantic1,ruan2019devil,zhao2021ordinal,xia2017joint,
fang2018weakly,liang2018look,xia2016zoom,nie2018mutual,li2021multi,wang2019learning,
gong2019graphonomy,liu2019braidnet,wang2020hierarchical,zhou2021differentiable,zeng2021neural,liu2021hierarchical} propose to address this critical issue using deep learning techniques, which build well-established parsing baselines for understanding human structures.

The MHP dataset~\cite{li2017multiple,zhao2020fine} is presented to address multiple human parsing challenges that involve multiple human identities in one image, in addition to the conventional human parsing tasks. Beyond this challenging task, CIHP~\cite{gong2018instance} is established to solve the instance-level human parsing task,~\eg~\cite{ruan2019devil,yang2019parsing,li2020self,he2020grapy,ji2020learning,zhang2020human,zhang2020part,loesch2021describe}, which not only requires the semantic information of foreground objects, but disentangles these parts into different human identities. Moreover, other works offer to investigate the part segmentation problem from a video-based perspective,~\ie, video-based human parsing, with the proposal of VIP~\cite{zhou2018adaptive}. Video-based part segmentation requires a semantic consistency of temporal sequences. In summary, the tasks of these new trends continue to be founded on the segmentation of human clothing, while ignoring human body structure exploration and leaving space for future research.

\textbf{Object part parsing.} Except for human parsing tasks, here we review the object part segmentation task and divide the existing literature into two different lines: 1) rigid object part segmentation~\cite{song2017embedding,car2014parsing,liu2022learning}: including cars, aeroplanes, motorbikes, and other vehicles; 2) non-rigid object part segmentation~\cite{wang2015semantic,wang2015joint,naha2021part}: including birds, horses, cows, and other living creatures. Although these two lines of segmentation tasks can be uniformly solved by the popular deep learning schemes, challenges remain due to ambiguous semantic representations, blurriness around part boundaries, and anti-topological predictions. However, as the rigid objects usually consist of stable structures, hence Song~\etal~\cite{song2017embedding} and Liu~\etal~\cite{liu2022learning} proposed to embed the canonical 3D models into learnable 2D part segmentation tasks, while in~\cite{car2014parsing}, the static geometric relationships are calculated for deducting the related part regions. However, when extending these strategies to non-rigid objects, namely articulated objects, the connections between object parts show a significant variance because of the various part shapes. Hence the dynamic part relationships~\cite{wang2015semantic,wang2015joint} or pose-aided strategies~\cite{naha2021part} are proposed to solve this problem, which will be elaborated on in the next subsection.

\textbf{Weakly-supervised part parsing.} Several recent works have proposed exploiting weak supervision to generate dense semantic part segmentation masks. Unlike conventional weakly-supervised semantic segmentation methods, part segmentation using conventional image-level or box-level supervision would inevitably lead to highly repetitive semantic meanings (~\eg, every human image contains the semantic label of \textit{torso} and \textit{legs}) and ambiguous annotations (bounding box overlap) respectively. Thus several recent works propose to use human pose information as a weak supervision, which also shows high relationships with the part segmentation masks. We would like to elaborate on this in~\secref{sec:strategy-part-deep}. Wu~\etal~\cite{wu2019keypoint} and Yang~\etal~\cite{yang2021weakly} propose to generate accurate part segmentation masks using keypoint annotations.
Moreover, Zhao~\etal~\cite{zhao2022pose} propose a pose-to-part framework that gradually transfers weak pose annotations to the accurate segmentation masks and then use the image-level boundaries to correct the ambiguous regions.

\textbf{Unsupervised learning.} The aforementioned part segmentation tasks require accurate pixel-level annotations. It is an extremely labor-consuming work~\cite{yang2021learning}, especially performing annotations in the fine-grained part levels. Hence another trend of works~\cite{gonzalez2018semantic,lorenz2019unsupervised,hung2019scops,gao2021unsupervised,
liu2021unsupervised,choudhury2021unsupervised} proposes to explore the semantic information through unsupervised manners. In~\cite{gonzalez2018semantic}, the automatic discovery of semantic parts and the relationships between the linguistical definition and activations discovered by CNNs are first explored. Leading by this thought, several works~\cite{lorenz2019unsupervised,hung2019scops} are proposed to leverage the advantages of deep representations. One specific feature is that the part representation after geometric transformation should be invariant over all instances of a category. Beyond this idea, Choudhury~\etal~\cite{choudhury2021unsupervised} propose to discover the object part by using the contrastive loss among local regions. In addition, as all object instances from one category share the same part compositions, Gao~\etal~\cite{gao2021unsupervised} propose to leverage the consistency of specific parts from different object instances, for example, different \textit{wings} of \textit{birds} share similar shapes and localization with respect to holistic objects.

The above works undoubtedly demonstrate the strong ability of automatic part discovery by adding constraints to deep neural networks. Without any prior guidance, these localized parts show a strong relation to the morphological structures of object categories. Thus a natural concern arises: can we use this object compositional information to guide the learning of other tasks? Furthermore, with weak supervision (\eg, image-level class labels), can we localize the accurate part regions that are most helpful for the learning objective, and what are the relations among these parts in recognition tasks? Keeping these concerns in mind, we will explain and discuss these details in~\secref{sec:fgvc}.

\begin{figure}[!t]
	\centering
	\includegraphics[width=1\linewidth]{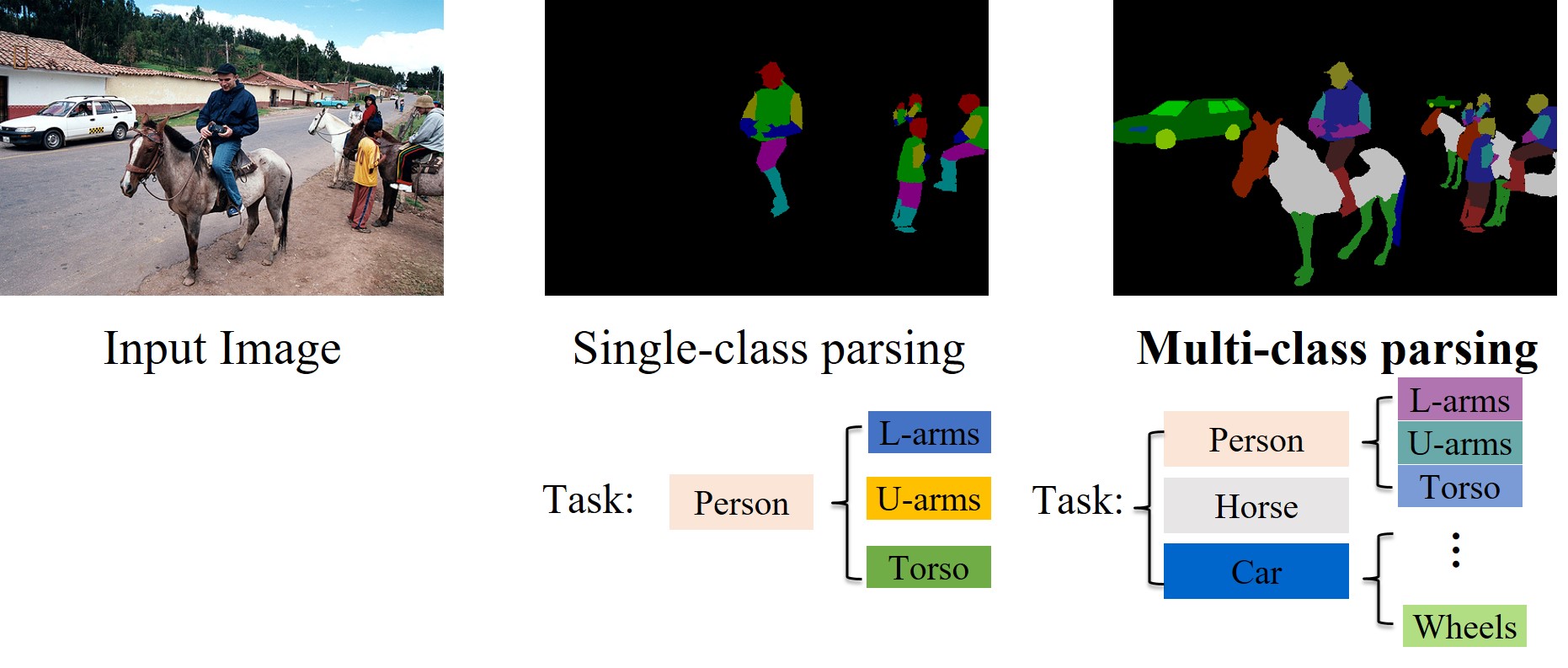}
	\caption{Task settings of single-class and multi-class part segmentation. The single-class part segmentation only focuses on segmenting the objects of one specific class, while multi-class part segmentation aims to segment multiple objects that occurred in one scenario.
	}
	\label{fig:multi-class}
\end{figure}

\subsubsection{Multi-class Part Segmentation}
When revisiting the single-class part segmentation task in~\figref{fig:multi-class} b), there remain challenging problems for understanding the image content. Focusing on a single class,~\eg, person class, and ignoring other meaningful classes,~\eg, cars, and horses, would lead to severe parsing issues for understanding the context. In~\figref{fig:multi-class}, only parsing human bodies into parts leads to a lack of object interaction with the context, for example, what is the human doing and where is the human sitting?

\textbf{Multi-class semantic part segmentation.} In response to these above challenges, the multi-class part segmentation tasks~\cite{zhao2019multi} are naturally proposed, as in~\figref{fig:multi-class} c). The multi-class part segmentation tasks aim to segment objects of multiple classes into parts. In~\cite{zhao2019multi}, a re-organized benchmark of the PASCAL-Part dataset is first proposed to solve this task, resulting in 58 semantic part classes. This new benchmark setting introduces additional challenges compared to the conventional single-class setting: 1) semantic ambiguity: parts of different object categories could share similar appearances,~\eg, horse and cow legs; 2) boundary ambiguity: part boundaries of different objects are usually hard to disentangle. Toward this end, pioneer work~\cite{zhao2019multi} proposes a joint boundary-semantic awareness framework with auxiliary supervision. In~\cite{michieli2020gmnet}, a graph-based matching network is proposed to construct the complex relationships between different parts, handling the part-level ambiguity and localization problems, which achieves success in handling part segmentation tasks of large scales,~\ie, 108 part classes.
While its journal version~\cite{michieli2022edge} focuses on the improvements of edge localization and extends the ADE-20k dataset~\cite{zhou2017scene} with part parsing labels, namely ADE 20K-Part.
Besides, Tan~\etal~\cite{tan2021confident} propose a semantic ranking loss to re-rank these semantic parts by their predicted confidence. Singh~\etal~\cite{singh2022float} develop a new learning framework that increases scalability and reduces task complexity compared to the monolithic label space counterpart. Additionally, this new research~\cite{singh2022float} introduces more complex part challenges,~\ie, distinguishing \textit{left} and \textit{right} part localizations with more than 200 semantic classes.

\textbf{Panoptic part segmentation.} Motivated by the Panoptic Segmentation proposed by~\cite{kirillov2019panoptic}, parsing objects into disjoint parts along with the background regions seems to construct a comprehensive interaction with the environmental context. Geus~\etal~\cite{de2021part} establish the Part-aware Panoptic Segmentation (PPS) task to understand a scene at multiple levels of abstraction. This PPS benchmark is founded on two representative datasets, PASCAL-VOC~\cite{Everingham15} for daily images and Cityscapes~\cite{Cordts2016Cityscapes} for autonomous driving. In~\cite{de2021part}, a two-stage semantic parsing framework is proposed and the evaluation criteria for this new task are founded.

\subsection{Strategies in Part segmentation}\label{sec:strategy-part}

Beyond the specific challenges of task setting in~\secref{sec:task-part}, part segmentation methods are designed following certain basic principles. Even the deep learning models and the non-deep ones share similar thoughts for constraining the optimization process. In this subsection, we will first introduce their commonalities and contrasts and then discuss and explore the promising future directions.

\subsubsection{Non-deep learning models: hand-crafted priors}
Object part parsing in the past decade does not strictly follow the current definition of semantic segmentation, but decomposing a holistic object into basic compositional units shares the same concerns. In~\cite{felzenszwalb2010object}, the Deformable Part Models are proposed to localize and understand the whole object, which constructs a feature pyramid with respective deformative locations. Eslami~\etal~\cite{eslami2012generative} propose a generative model to jointly learn the appearances and part shapes and use block-Gibbs Markov Chain Monte Carlo (MCMC) for fast inference.
Following this trend, Liu~\etal~\cite{liu2014fashion} adopt the Markov random fields to model the color and appearance similarities, deciding the part belongings. Meng~\etal~\cite{meng2017seeds} propose to initialize part seed proposals and then develop a seed propagation strategy to combine other potential regions. Some other researches~\cite{desai2012detecting,azizpour2012object,zhang2014part} segment object parts as an intermediate result to help the downstream tasks, including object detection, pose estimation, and action recognition.

Besides these works, the other line of works proposed to build trees~\cite{zhu2008unsupervised,hsieh2010segmentation,wang2011learning,dong2015parsing,xia2016pose} or graph models~\cite{car2014parsing,chen2014detect,wang2015semantic}, depicting the \textit{relationships of different object parts}. In~\cite{zhu2008unsupervised}, a joint bottom-up and top-down procedure is proposed to hierarchically decompose the holistic object into coarse parts, fine-grained parts, and basic lines/keypoints. Wang~\etal~\cite{wang2011learning} introduce hierarchical poselets, which decompose the human bodies into poselets (\eg, torso + left arm). Moreover, Several studies~\cite{dong2015parsing,xia2016pose} construct ``And-Or" graphs to assemble the outputs of parts.~\eg, Dong~\etal~\cite{dong2015parsing} build a deformable mixture parsing model to simultaneously handle the deformation and multi-modalities of Parselets.
Other works resort to graph structures, which are relatively flexible compared to hierarchical trees. For example, Chen~\etal~\cite{chen2014detect} construct a relational graph by the part attributes itself and pair-wise relationships. Wang~\etal~\cite{wang2015semantic} propose to learn the part compositional model under multiple viewpoints and poses, constructing a robust transformation of different conditions.

\textbf{Revisiting non-deep learning models.} With the development of deep learning techniques~\cite{he2016deep,simonyan2014very,he2015delving}, there is no doubt that the deep part segmentation models occupy the predominant places, benefited from their significant leading performance.
Following the end-to-end training framework~\cite{long2015fully} in semantic segmentation, recent part segmentation models achieve more success than the conventional hand-crafted feature extractors,~\eg, HOG or SIFT features. However, these deep learning models neglect the consideration of hierarchical body structures and would face great challenges in understanding unseen data and generating unreasonable segmentation masks. For example, in human parsing tasks, deep models always follow the statistical rules that deservedly take the round-shaped objects as human heads, which leads to incorrect parsing results for car wheels. In some error parsing cases, the lower legs could be connected with the upper bodies which breaks the basic topological rules. Interestingly, these phenomena are usually rare in conventional non-deep learning models, which follow the strict constraints of topological or morphological compositional principles,~\eg, the human bodies are hierarchically decomposed into basic structures thus adjacent body parts show strong correlations.  Moreover, the non-deep learning models require very little training data, showing great application potential in handling extreme circumstances in real-world applications.

\subsubsection{Deep learning strategies}\label{sec:strategy-part-deep}
In addition to these aforementioned differences, the non-deep learning and deep learning models share similar designs and basic foundations to solve the fine-grained part parsing tasks. Whether hand-crafted feature extractors or deep feature extractors are employed, the basic challenges still remain for parsing reasonable and clear segmentation results. Three important characteristics of deep learning-based models are discussed in this subsection.

\textbf{Pose-aided learning.} The pose estimation and part segmentation are dual problems. Compared to the dense pixel prediction task of part segmentation, pose estimation is a more lightweight estimation task with significantly less annotation consumption. Conventional non-deep learning methods~\cite{yang2011articulated,yamaguchi2012parsing,dong2014towards,xia2016pose} have proposed the importance of joint learning of these two related tasks. In the era of deep learning, major research efforts~\cite{xia2017joint,liang2018look,nie2018mutual,fang2018weakly} focus on the joint optimization of human part parsing and pose estimation with the proposals of large datasets. In~\cite{nie2018mutual}, a mutual learning framework is proposed by embedding the dynamic kernel of pose estimation to part segmentations. Fang~\etal~\cite{fang2018weakly} propose to transfer the human pose estimation knowledge as a coarse parsing prior and then to refine these coarse masks in the subsequent stages. Besides, other weakly-supervised methods~\cite{zhao2022pose} using keypoints information also achieves notable successes.
Methods of this category verify that the accurate localizations of pose key points, including animals and human beings, shows strong benefits to the tasks of part segmentations.

\textbf{Multi-scale zooming.} Different from the object segmentation tasks, part segmentation demonstrates a great demand in parsing detailed regions inside objects. Chen~\etal~\cite{chen2018deeplab} propose the atrous convolutional network to enhance the receptive field while~\cite{chen2017rethinking} introduces an improved structure of atrous spatial pyramid pooling (ASPP), which incorporates multi-scale features in one single layer. Besides these general improvements, in~\cite{chen2016attention}, a two-stream CNN is proposed to fuse the global features and local features. While Xia~\etal~\cite{xia2016zoom} propose a stage-wise framework to detect and segment object parts from image-levels to object-levels and then part-levels.

\textbf{Part relationship guidance.} Several recent works~\cite{liang2016semantic,zhao2021ordinal,wang2015joint,gong2019graphonomy,wang2020hierarchical,zhang2020part,he2020grapy,wang2019learning} propose to embed the part-level relationship as learning priors to guide the segmentation process. For example, Wang~\etal~\cite{wang2015joint} propose a joint CRF to model the object-part and part-level relationships after the encoding of image features.
\cite{zhao2021ordinal} decouples the part segmentation learning as multiple independent tasks while using the part-level learning order to constrain the recurrent learning process. Gong~\etal~\cite{gong2019graphonomy} adopt a universal graph learning strategy to model the part relationship across multiple datasets. Wang~\etal~\cite{wang2020hierarchical} propose a hierarchical part parsing network to gradually decompose the object from the coarse level to the finer level.
In addition, in~\cite{wang2019learning}, a tree structure is constructed based on the CNN architectures and models the part-level relationship for understanding. Methods of this category successfully incorporate relationship learning to promote the segmentation process, while also using the accurate feature extraction of CNNs.  The key challenge in fine-grained visual parsing is to understand the compositional relationships. Here we summarize these relationships as follows: 1) object-part relationship; 2) part-level relationship within one object; and 3) part-level across different images/objects.
By understanding these relationships, deep models can further promote the learning of action recognition, fine-grained object recognition, and re-identification tasks.

\section{Fine-grained Object Recognition: Understanding Local Structures }\label{sec:fgvc}
\subsection{Definition and Challenges}
\textbf{Definition.} Image Object classification has achieved great success benefiting from the development of deep learning systems and proposals of large datasets. Here we conclude the tasks of image object classification as \textit{base-level recognition}, for example, classifying horses and aeroplanes, as in~\figref{fig:motivation}. Objects in base-level categories can be easily distinguished by image-level global features and usually has large margins in semantic definitions. For fine-grained object recognition, deep learning systems are required to distinguish the subtle differences among sub-categories that have similar appearances and semantic definitions. In this problem, methods developed for \textit{base-level recognition} usually face great challenges for classifying \textit{fine-grained} classes as in~\figref{fig:motivation}.

The formulation of fine-grained object recognition is similar to the common base-level ones, by learning using a much more ``compact" semantic label space.
Moreover, the generalized definition of fine-grained object recognition problems consists of two different levels of recognition: 1) subcategory-level: recognizing different fine-grained sub-categories that consist of multiple identities; 2) instance-level: distinguishing and identifying two different instances,~\eg, person re-identification, vehicle re-identification, and face recognition. In this survey, we mainly focus on the first level of research but it should be noted that these two sub-field share many common techniques, which will be discussed in the following section.

\begin{figure}[!t]
	\centering
	\includegraphics[width=1\linewidth]{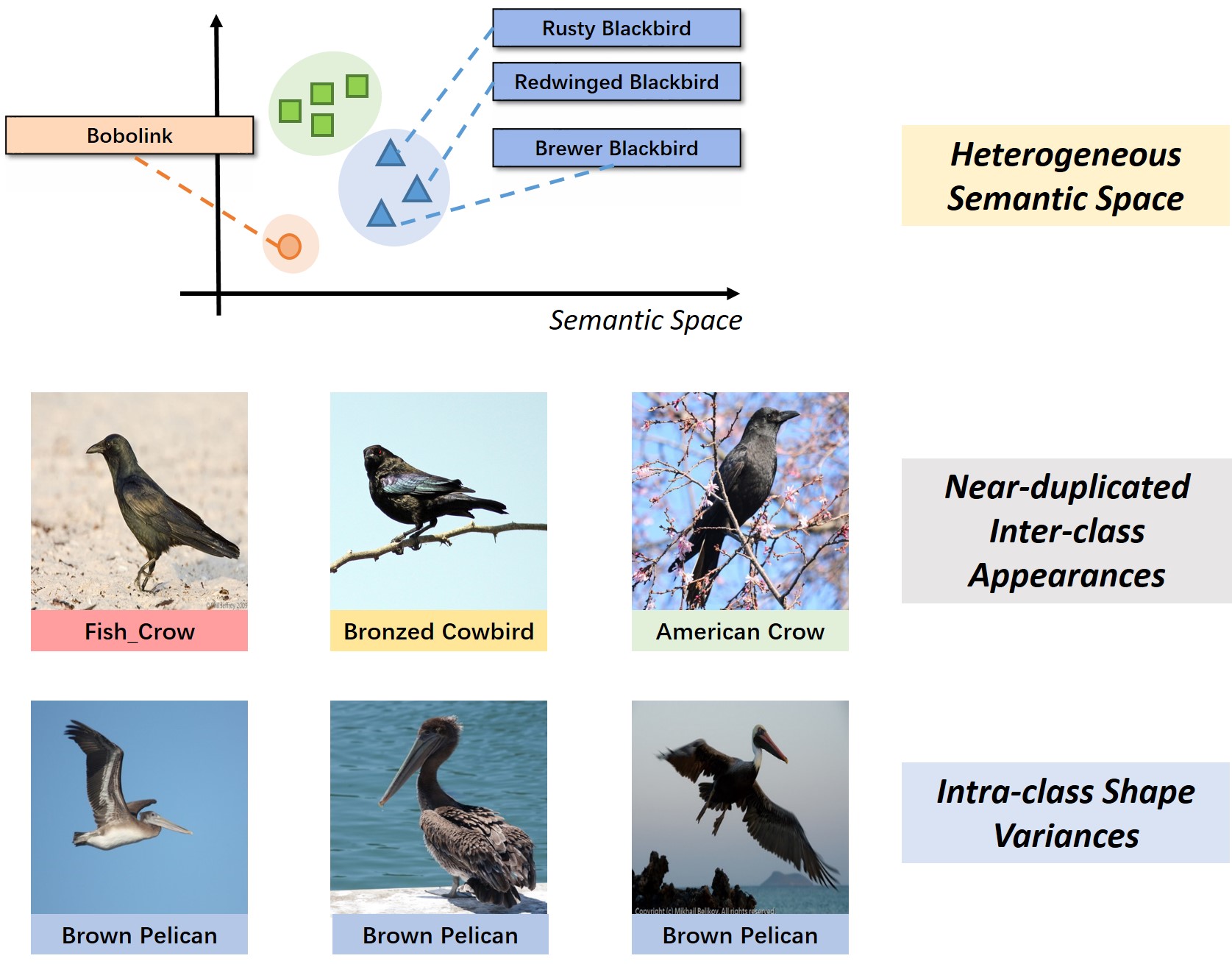}
	\caption{Three typical challenges in fine-grained recognition tasks (images from CUB dataset~\cite{wah2011caltech}). 1) Heterogeneous semantic spaces: the semantic definitions of fine-grained text labels are usually cluster distributed. 2) Near-duplicated inter-class appearances: objects of different categories present visually similar appearances. 3) Inter-class shape variances: the shape structures of objects in the same category can be inconsistent.
	}
	\label{fig:challenge}
\end{figure}

\begin{table*}[t]
\caption{Summarization and comparisons of 13 widely-used part segmentation benchmark datasets. The Bbox and Part in annotation items indicate that the dataset provides object bounding box labels and part-level localization labels respectively. }\label{tab:fgvc-dataset}
\centering
\renewcommand{\arraystretch}{1.4}
\resizebox{\linewidth}{!}{
\begin{tabular}{l|c|c|c|c|c|c}
\toprule
Dataset & Pub. &Year  & Image Num. & Category & Annotation & Description\\
\midrule
Oxford 102 Flowers~\cite{nilsback2008automated}&ICCVGIP&2008& 8,189& 102& - & Flower Classification\\
CUB-200-2011~\cite{wah2011caltech}&-&2011&11,788 & 200& Bbox \& Part & Birds, Best-known fine-grained benchmark\\
Stanford Dogs~\cite{khosla2011novel}&CVPRW&2011& 20,580&120& Bbox & Dog Classification \\
Stanford Cars~\cite{krause20133d}&ICCVW&2013& 16,185 &196 & Bbox &  Car Classification\\
FGVC Aircraft~\cite{maji2013fine}&Arxiv&2013& 10,000&100& Bbox & Aircraft Classification\\
Food 101~\cite{bossard2014food}&ECCV&2014 &101,000& 101 &- & Food Classification\\
BirdSnap~\cite{berg2014birdsnap}&CVPR&2014& 49,829& 500 &Bbox \& Part& Large Bird datasets \\
NAbirds~\cite{van2015building}&CVPR&2015& 48,562 &555 &Bbox \& Part & Large Bird datasets\\
CompCars~\cite{yang2015large}&ECCV&2018&136,727&431& Part Images&Cars from web-nature and surveillance-nature\\
DeepFashion~\cite{liu2016deepfashion}&CVPR&2016& 800,000 & 1,050&Bbox \& Part & Clothes Classification\\
iNat2017~\cite{van2018inaturalist}&CVPR&2018&857,877& 5,089 &Bbox &Large-scale Species Classification\\
Dogs-in-the-Wild~\cite{sun2018multi}&ECCV&2018&299,458&362&-&Large-scale Dog Classitication\\
iNat2021~\cite{van2021benchmarking}&CVPR&2021&3,286,843 & 10,000&-&Improved version of iNat2017~\cite{van2018inaturalist} \\
\bottomrule
\end{tabular}
}
\end{table*}

\textbf{Challenges.} Here we summarize three typical challenges of the fine-grained object recognition task in~\figref{fig:challenge}: 1) Heterogeneous semantic space: although fine-grained labels are distributed in a compact space compared to the base-level category labels, their semantic definitions are still heterogeneous. For example, there are three types of \textit{blackbirds} but only one \textit{bobolink} in the semantic space. This phenomenon is still less-explored in the field of fine-grained recognition which leaves challenges for learning appropriate decision boundaries. 2) Near-duplicated inter-class appearances: in the middle of~\figref{fig:challenge}, we present three images from different fine-grained categories while sharing much common ground in visual appearances. Thus, deep learning models need to clearly distinguish their differences by observing local details. 3) Intra-class shape variances: image objects that belong to the same categories can present in various shapes and structures. As for the bird classification task, the flying attitude shares less intuitive visual cues with that sitting one, bringing challenges to deducting these images with various shapes in the same categories.
In most cases, these three challenges show mutual effects on each other, and a good learning model needs to have the ability to handle semantic imbalance, inter-class similarities, and intra-class diversities simultaneously.

\subsection{Benchmark Datasets}
In this subsection, we summarize the prevailing benchmark datasets in the field of fine-grained recognition. In~\tabref{tab:fgvc-dataset}, with the development of machine learning systems, earlier works have established benchmark datasets with more than 100 categories for the classification of common daily objects, including Oxford 102 Flowers~\cite{nilsback2008automated} for plants, CUB-200-2011~\cite{wah2011caltech} for more than 200 bird categories, and Stanford-Dogs~\cite{khosla2011novel} for the classification of 120 dog sub-categories.

Pioneer machine learning methods, including SVM, and dictionary learning face great challenges in tackling these problems with less than 30\% accuracies~\cite{zhang2012pose}, indicating that these works cannot be directly used in real-world industrial applications. Thus to solve these problems, these datasets provide bounding box information for localizing the main objects and providing the box or segmentation masks for part learning,~\eg, bird heads and torsos for the CUB dataset~\cite{khosla2011novel}. Integrating this fore-ground information or part localization priors significantly helps the learning of fine-grained objects, especially the subtle differences of near-duplicated objects.

With the development of deep learning systems, especially CNNs, the representation ability for fine-grained objects has been significantly improved, \eg, from 28\% to 75\% accuracy on the CUB-200-2011 benchmark in~\cite{zhang2014part}. Despite their effectiveness, deep learning models usually rely on the acquisition of a large number of training data with similar distributions. Thus many new datasets with plentiful annotations are proposed, including Food 101~\cite{bossard2014food} with more than 101k images, NAbirds~\cite{van2015building} and BirdSnap~\cite{berg2014birdsnap} for nearly 50k images.
These datasets not only provide high-quality annotations but also introduce new challenges for complicated semantic definitions, intra-class diversities, and inter-class similarities.

Beyond these earlier deep learning benchmarks, recent advanced research proposes large-scale annotations with a huge number of fine-grained categories. For example, iNat2017~\cite{van2018inaturalist} provides more than 0.85M images of 5k categories, while its improved version iNat2021~\cite{van2021benchmarking} provides more than 3M images of 10k categories.
In addition, several other large-scale benchmarks~\cite{zhuang2018wildfish,weyand2020google} have been proposed for fish recognition and landmark recognition.
Beyond the aforementioned challenges, these datasets span the new dilemmas: 1) imbalanced/long-tailed data distributions: objects of some rare categories usually consist of few annotations, while other main classes consist of thousands of images; 2) noisy labels: images of large scale datasets are usually collected webly and would introduce many noisy ambiguous labels. Thus the classification model needs to further purify these noisy factors by learning from predominant clean annotations.

\subsection{Strategies in Fine-grained Recognition}
Recognizing fine-grained objects has attracted much research attention in the last two decades. In this subsection, we first conclude the non-deep learning techniques including the hand-crafted features and human-in-loop learning frameworks in~\secref{sec:nondeepfgvc}. We then conclude the recent advanced research using deep learning techniques from two aspects,~\ie, the part-guided learning in~\secref{sec:part-guidedfgvc} and learning with feature representation constraints in~\secref{sec:featurefgvc}.

\subsubsection{Non-deep learning Models}\label{sec:nondeepfgvc}

\textbf{Hand-crafted feature extraction.} Pioneer fine-grained works~\cite{yao2011combining,yao2012codebook,zhang2012pose,goring2014nonparametric} propose to use hand-crafted features to recognize objects. For example, Zhang~\etal~\cite{zhang2012pose} propose to incorporate the SVM into understanding pose structures, learning with SIFT and BoW (Bag of Words) features. Yao~\etal~\cite{yao2011combining} propose dense sampling strategies with random forests to extract local features. Other researches including~\cite{yao2012codebook} adopt the codebook learning strategy for encoded dictionaries. Methods of this category can benefit from learning with local descriptors or part features while still facing difficulties in understanding fine-grained semantics.

\textbf{Human in the loop.} Conventional machine learning methods usually lead to relatively low performance,~\eg, 28\% for CUB-200-2011 classification tasks, which have difficulties for applying in realistic applications. Hence earlier works propose to incorporate human expert knowledge into the learning process. For example, Wah~\etal~\cite{wah2011multiclass} leverage computer vision techniques and analyze the user responses to gradually enhance the final learning accuracies. While Branson~\etal~\cite{branson2011strong} propose an interactive scheme with deformable part models to distinguish the subtle differences between similar objects.

\subsubsection{Part-guided Learning}\label{sec:part-guidedfgvc}

\textbf{Supervised part learning.} With the development of deep learning techniques, recognizing common base-level objects has made significant progress. Although the performance of fine-grained recognition has been improved in many applications~\cite{zhang2014part}, considering the challenges mentioned above, distinguishing subtle differences among near-duplicated objects usually faces serious dilemmas. Thus dozens of works propose to employ the part-level features~\cite{zhang2014part,huang2016part,he2017weakly,wei2018mask,he2019part,
peng2017object,huang2020interpretable,wang2015multiple} to amplify these differences in a local perspective. Zhang~\etal~\cite{zhang2014part} propose a part-based R-CNN to locate the part features and then build pose-normalized features as the enhancement for global features. Krause~\etal~\cite{krause2015fine} propose to use keypoint annotations for fine-grained recognition, which leverages the co-segmentation techniques to align different views of images. Huang~\etal~\cite{huang2016part} propose a dual-stream part-stacked CNN to jointly learn discriminative features from high-resolution part features and low-resolution global ones. In addition to these techniques using part detection methods, Wei~\etal~\cite{wei2018mask} propose to use part segmentation masks to regularize the local descriptor learning process. Although segmentation masks provide more accurate learning guidance, learning with all feasible part proposals would lead to a globally homogeneous amplification of every pixel.
To summarize, supervised part-learning techniques adopt the part detectors or segmentation masks as local feature selection guidance, and then fuse these local features with the image-level global ones. In this manner, not only the global features but the local details are taken into account for final feature distance measurements. However, these part excavation methods still rely on accurate part segmentation or detection annotations, requiring enormous labor consumption. In addition, considering that the accurate part annotations of test data are usually infeasible, transferring this expert part knowledge into the testing environment would also lead to an inductive bias, which would further limit the effectiveness in real-world applications.

\begin{figure*}[!t]
	\centering
	\includegraphics[width=1\linewidth]{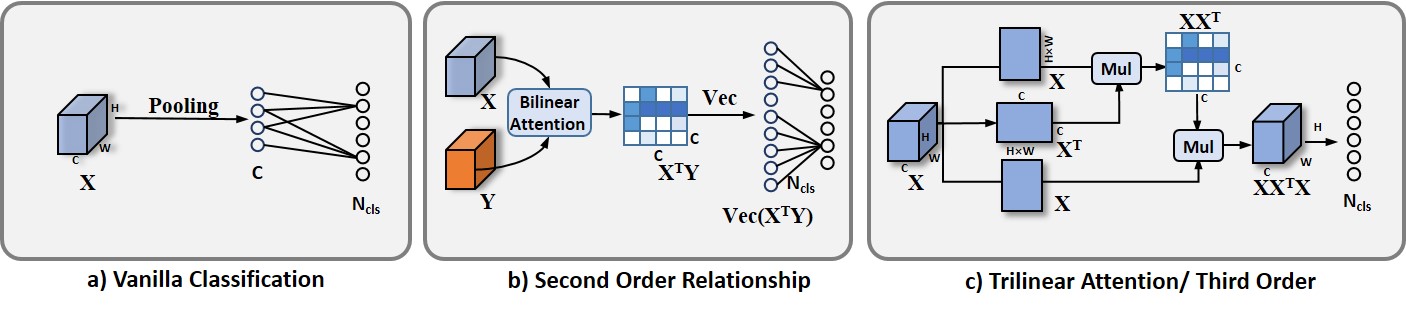}
	\caption{Three typical high-order relations as in~\cite{zhao2021graph}. Vanilla classification: encoded features are pooled into vectors for classification, used in most of the works. Second-order relationship~\cite{lin2015bilinear,gao2016compact,li2017factorized,wei2018grassmann,zhao2021graph}: learning rich second-order features by keeping the spatial dimension. Trilinear attention~\cite{zheng2019looking,gao2020channel,wang2018non}: preserving the same size as input features for learning spatial-wise or channel-wise attention matrix.
	}
	\label{fig:ordercomp}
\end{figure*}

\textbf{Unsupervised part learning.}\footnote{also noted as weakly-supervised fine-grained recognition.}
Considering the labour-intensive computation and unstable generalization ability in the inference stage, recent ideas~\cite{simon2015neural,zhang2016weakly,fu2017look,recasens2018learning,wang2020weakly,ge2019weakly,sun2020fine,
zheng2019learning,ding2019selective,wang2020graph} propose to use unsupervised part attention techniques. In~\cite{gonzalez2018semantic}, authors prove that during the back-propagation process, neural networks have the potential to discover semantic parts automatically. Leading by this trend, Simon~\etal~\cite{simon2015neural} propose neural activation constellations to localize semantic parts without any supervision. Different from this work, Fu~\etal~\cite{fu2017look} propose a multi-stage zooming strategy to automatically locate and re-scale the attention regions, by learning the confidence scores of different zooming proposals. Similar to this work, Recasens~\etal~\cite{recasens2018learning} develop a saliency-based sampling layer for neural networks after finding the activated regions. While Ge~\etal~\cite{ge2019weakly} incorporate the weakly-supervised detection and segmentation models for localizing the discriminative features for fine-grained distinguishing. Besides, Wang~\etal~\cite{wang2020weakly} propose a Gaussian mixture model for investigating the object parts with an auxiliary branch for supervision.  In~\cite{wang2020graph}, a graph-propagation correlation learning method is proposed to model and propagate the discriminative part features to other parts. Nevertheless, these methods have shortcomings in two aspects: 1) introducing auxiliary learning branches or stages for optimization; and 2) the number of part proposals can sometimes be large whereas only a few are useful for recognition.

To solve this issue as well as reduce computational costs, Lam~\etal~\cite{lam2017fine} propose an HSNet searching architecture to explore the most discriminative parts, while other work~\cite{he2017weakly} builds a weakly-supervised part selection mechanism based on their response scores. Zhao~\etal~\cite{zhao2021part} propose a Transformer architecture to build inter-part relationships and adopt multiple auxiliary branches for part-awareness learning, while in the inference stage, these auxiliary branches are not used for computational consideration.
Methods of unsupervised part learning~\cite{ji2020attention,nauta2021neural,zhao2021part,yang2018learning} lead the prevailing trend in fine-grained recognition, which benefits from its strong ability in understanding local differences and discovering object parts. Furthermore, selecting and modeling the part relationships becomes an emerging topic in fine-grained recognition.

Different from the unsupervised part learning in semantic segmentation, part attention in fine-grained recognition aims to discover the discriminative features and exploits these local features as an enhancement for distinguishing near-duplicated objects. Thus the semantic information of the unsupervised part in recognition is usually not strictly aligned with the natural common definitions.

\subsubsection{Feature Representation Learning}\label{sec:featurefgvc}
Besides the methods using part-level features for enhancing the local details, the other crucial problem in fine-grained recognition is feature representation learning. There is intuitive thinking that well-represented features can provide a more robust and generalization ability for downstream tasks, including segmentation, detection, and also fine-grained recognition. Despite the experimental evidence, enhancing the detailed representation ability helps the measurement of local subtle differences hidden among different features, which may be vital factors for discrimination. When features of various images are distributed in one generalized and robust fine-grained feature space, these subtle differences would be easy to discover. Guided by the theory, this line of methods tends to regularize the feature learning process~\cite{wang2018learning,dubey2018pairwise,aodha19presence,dubey2018maximum,cui2018large,zheng2019towards} or generate rich feature representations~\cite{lin2015bilinear,gao2016compact,li2017factorized,wei2018grassmann,wei2017selective,
yu2018hierarchical,zheng2019looking,gao2020channel,zhang2019learning,zhao2021graph} without using additional annotations.

\textbf{High-order representations.}  As in~\figref{fig:ordercomp} a), given an input image $\mc{I}$, the conventional classification model can be represented as  $\mathbf{X} = \Phi(\mc{I})$. Thus $\mathbf{X} \in \mathbb{R}^{W \times H \times C}$ denotes the $C$-dimensional with $H\times W$ feature maps and the final classification vector would be $\bs{Pool}(\mathbf{X}) \in \mathbb{R}^{1 \times 1 \times C}$. Considering the spatial feature relationship is neglected during the pooling operations, high-order interactions are proposed in advanced works. In~\figref{fig:ordercomp} b), as the pioneer work using the second-order relationship, Lin~\etal~\cite{lin2015bilinear} propose a bilinear model to extract shape and appearances by two different CNNs and then construct a bilinear pooling operation to generate rich second-order representations,~\ie, $\mathbf{X}= \frac{1}{WH} \sum_{i=1}^{W}\sum_{j=1}^{H}  \bs{vec}(\Phi_{1}(\mc{I})^{\top}_{i,j} \Phi_{2}(\mc{I})_{i,j})$. Although this bilinear pooling operation enriches the fine-grained representation and amplifies the differences of similar embedding, it also introduces high computational costs.~\ie, $C\times C\times N_{cls}$ for optimization, and $N_{cls}$ denotes the category numbers. To solve this,
Gao~\etal~\cite{gao2016compact} propose a compact bilinear pooling model that uses the network itself to build second-order relationships,~\ie, $\mathbf{X}\equiv\mathbf{Y}$. Other works propose to use matrix factorization~\cite{li2017factorized}, Grassmann constraints~\cite{wei2018grassmann}, and low-rank learning~\cite{kong2017low} to reduce computation costs. Besides these works, Yu~\etal~\cite{yu2018hierarchical} propose a hierarchical feature interaction operation to build heterogeneous second-order relationships. Zhao~\etal~\cite{zhao2021graph} propose a graph-based high-order relationship learning to reduce the high-dimension space into discriminative low dimensions.

However, the second-order feature learning still introduces the curse of dimensionality for optimization, thus the other line of works proposes to use the third-order relationship in~\figref{fig:ordercomp} c), namely trilinear attention~\cite{zheng2019looking,gao2020channel} or non-local mechanisms~\cite{wang2018non}. For example, Zheng~\etal~\cite{zheng2019looking} propose the trilinear attention in the channel-dimension with a distillation mechanism, which can be formulated as: $\bs{softmax}(\mathbf{X}^{\top}\mathbf{X})\mathbf{X}^{\top} \in \mathbb{R}^{WH\times C}$. While Gao~\etal~\cite{gao2020channel} propose a contrastive loss to learn the channel-wise relationship of inter-and intra-images. Methods using the third-order relationship maintain the output size and can be embedded into different network stages to enhance the representations.

\begin{figure*}[!t]
	\centering
	\includegraphics[width=\linewidth]{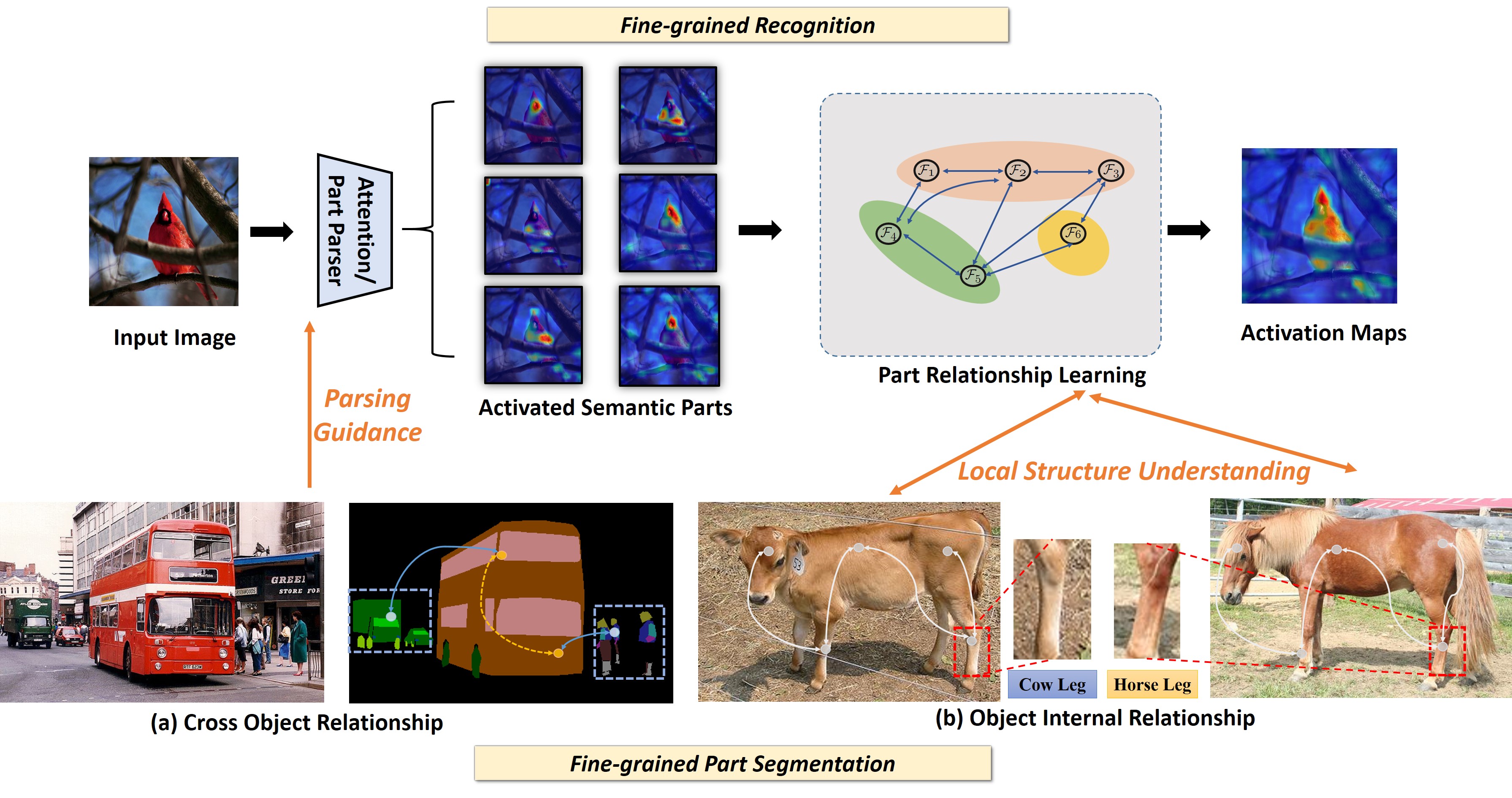}
	\caption{ Part relationship learning in two representative fine-grained visual tasks, fine-grained part segmentation~\cite{zhao2019multi} and fine-grained recognition~\cite{zhao2021graph}. First row: understanding complex fine-grained images requires the accurate parsing of local part relationships.
Second row: the cross-object relationship uses contextual information to help the understanding of small parts, while the object internal relationship with other parts helps the distinguishing of locally similar regions. Besides, the segmentation results can serve as parsing guidance and relationship learning in both tasks constructing the robust local structure understanding.
	}
	\label{fig:relation-example}
\end{figure*}

\textbf{Feature interactions and regularization.} Besides building high-order rich features, the other line of works proposes to use feature interactions~\cite{wang2018learning,sun2018multi,chen2019destruction,luo2019cross,zhuang2020learning,
liu2019bidirectional,rodriguez2018attend,liu2020filtration} or using additional constraints~\cite{wu2021align,wang2022bridge,dubey2018pairwise,aodha19presence,dubey2018maximum,chang2021your}. Wang~\etal~\cite{wang2018learning} construct a discriminative feature bank of convolutional filters that captures class-specific discriminative patches.  Sun~\etal~\cite{sun2018multi} propose a multi-attention multi-constraint network to regularize the feature distributions based on the selected anchors. While Luo~\etal~\cite{luo2019cross} propose to learn cross-level and cross-images relationships for building interactive feature representations. For robust feature learning, Chen~\etal~\cite{chen2019destruction} incorporate an additional destruction and construction branch as an additional learning task.
These works rely on additional blocks or feature interaction networks, which may introduce additional computation costs.

Besides works using additional parameters to feature enhancement, several works propose to use auxiliary constraints in addition to the basic cross-entropy constraints. In~\cite{dubey2018pairwise}, pair-wise confusion is proposed among Siamese networks to alleviate the overfitting issues. While Dubey~\etal~\cite{dubey2018maximum} propose an entropy maximizing approach to regularize the final classification confidence. Aodha~\etal~\cite{aodha19presence} propose geographically guided loss functions that deduct the fine-grained features using temporal and geographical spatial priors.
Besides, other works~\cite{wu2021align,wang2022bridge,zhou2020look} follow a self-supervised learning trend for fine-grained recognition. Wu~\etal~\cite{wu2021align} propose to solve the dilemma between self-supervised learning and fine-grained recognition by enhancing the salient foreground regions.

To summarize, 1) methods using high-order relations modules mainly focus on rich representations at the feature-level, and enhancing these representations would amplify the subtle differences among different object features, and 2) methods using additional constraints make fine-grained features to be distributed in compact and precise spaces, while alleviating the overfitting issue and concentrating more on object regions. This overfitting issue is further studied in existing works by generating accurate class activation maps while preventing only focusing on the local part regions. In the next section, we will discuss why we need local details and why only local details cannot perform accurate fine-grained recognition.

\section{Part Relationship in Segmentation and Recognition}\label{sec:partrelation}
Fine-grained visual parsing, including recognition, segmentation, detection and other high-level image understanding tasks, leaves us with challenges in its detailed and complex ``fine-grained" parsing requirements. Understanding images with fine-grained objects can be substantially different from common ``coarse-grained" ones. In this section, we investigate two representative fine-grained visual tasks,~\ie, segmentation and recognition with the following natural concerns:
\begin{enumerate}
\item What are the key challenges in fine-grained recognition, or what are the unique problems in this subfield?
\item Why does part relationship learning help the understanding of these fine-grained tasks? What are the relations among them?
\end{enumerate}
\subsection{Problems in Fine-grained Parsing}\label{sec:problem-finegrained}
Fine-grained visual parsing is a relatively-defined concept compared to the common daily categories. Considering the specific tasks of fine-grained recognition and semantic part segmentation, understanding image objects would face the following challenges.

1) \textbf{Non-salient/less-prominent in image-level:} the fine-grained visual features are \textit{imperceptible} using existing learning systems or not easily understood by human visual systems. In fine-grained recognition, objects of different semantic classes usually share visually similar appearances but still show \textit{imperceptible} discrepancies. This means that objects of these categories are recognizable by detailed local differences while understanding them using coarse global features is impracticable.
In the task of fine-grained part segmentation, distinguishing different parts relies on subtle visual cues, including \textit{imperceptible} part boundaries and near-duplicated local visual patterns. The term \textit{imperceptible} here means that these part boundaries can be relatively non-significant compared to object silhouettes and even contextual noisy information. In addition, in some extreme cases, due to the small scale of object parts (in~\figref{fig:relation-example} a)), it is difficult to recognize them by only observing small objects themselves. And in some cases, these discriminative cues in both parsing tasks are relatively non-significant, and are suppressed in the image-level feature representations.

2) \textbf{Locally distinguishable and indistinguishable:} as mentioned in the first challenge, the fine-grained objects are only recognizable in local details,~\eg,~\figref{fig:challenge} in~\secref{sec:fgvc}. Thus enhancing the representation of these local parts is beneficial to learn discriminative embedding. However, these local part regions are not always distinguishable, for example, birds of two different categories may exhibit similar appearances in torso, tail, and wings while only differing in their heads. Analogous to the fine-grained recognition task, we present a similar segmentation problem in~\figref{fig:relation-example} b). The legs of the horse and cow are locally indistinguishable whereas the holistic object categories are easy to recognize and segment, since their head regions are salient for distinguishing. To sum up, the local regions of an image may be distinguishable or indistinguishable when compared with different images.

3) \textbf{Ambiguous semantic definition:} the last challenge in fine-grained parsing is the ambiguous semantic definition, which is less explored in recent works. Conventional part parsing works~\cite{zhu2008unsupervised,hsieh2010segmentation} propose to build hierarchical structures,~\eg, from the holistic body to object parts, and then to line segments. However, considering the goal of segmentation and recognition, the semantic definitions of ``fine-grained" tasks become an increasingly critical problem. For segmentation tasks, LIP dataset~\cite{liang2018look} defines the human bodies with different fashion clothes,~\eg, skirts and coats, while other datasets~\cite{chen2014detect} tend to segment bodies with the morphological rules,~\ie, upper and lower bodies. Similarly, if we define two different categories of objects, ~\eg, \textit{bulldogs} and \textit{poodles}, the \textit{bulldogs} can be subdivided into \textit{English bulldogs} and \textit{American bulldogs} with fewer differences. Thus the definitions of fine-grained semantics leave us with severe challenges, or in other words, ``how fine is fine-grained parsing"?

\begin{figure}[!t]
	\centering
	\includegraphics[width=\linewidth]{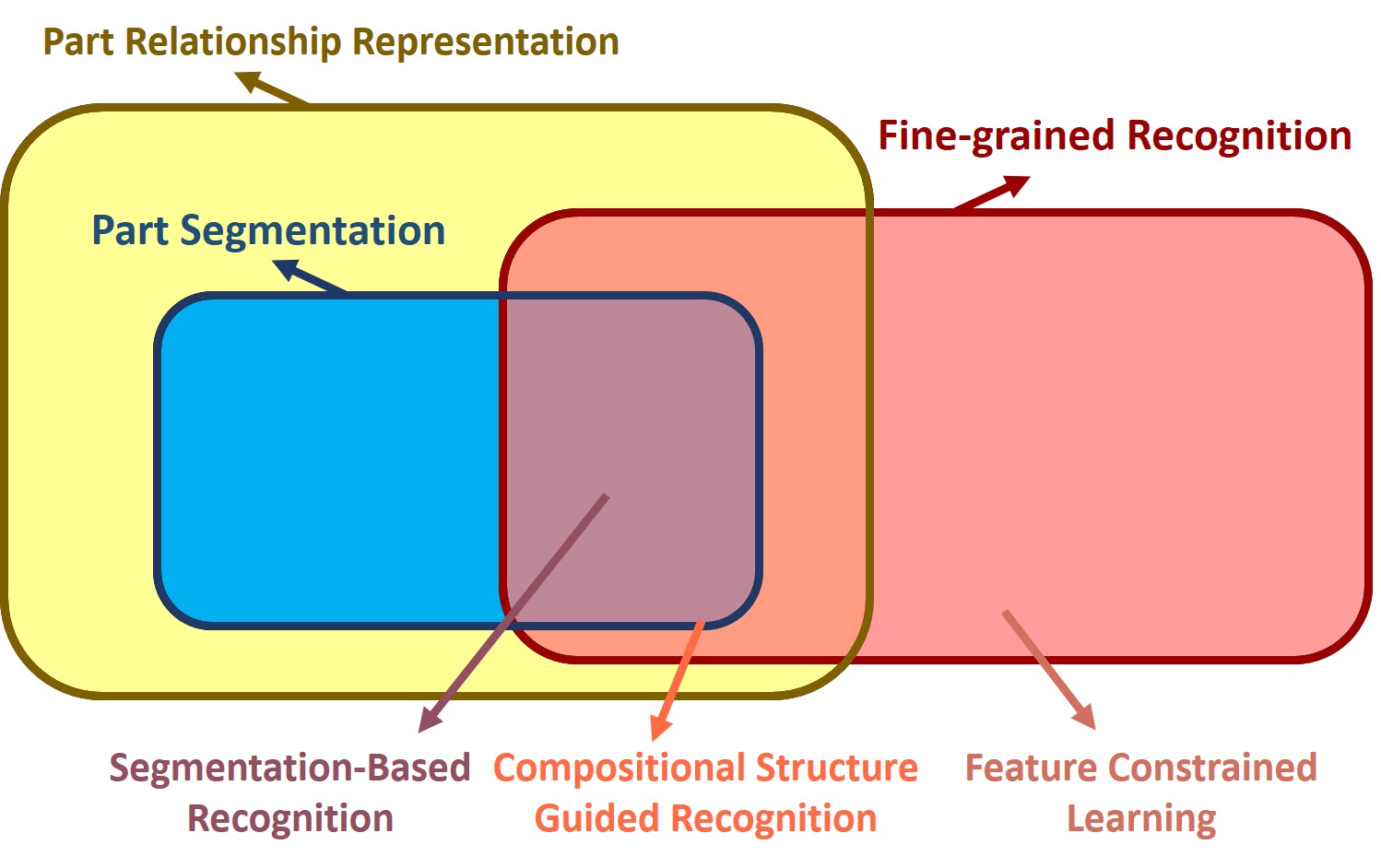}
	\caption{Correlations schematic of three relevant tasks,~\ie, part relationship representation, part segmentation and fine-grained recognition.
	}
	\label{fig:relations}
\end{figure}

\subsection{Part Relationship Learning: A Solution }\label{sec:part-relationsolution}
Towards the aforementioned challenges, in this survey, we argue that building \textit{Part Relationships} in fine-grained visual parsing would be one reliable and promising solution. Here we elaborate on its effectiveness in solving these challenges: 1) considering that the fine-grained cues are usually hidden in local details and cannot be distinguished using the image-level features. Hence enhancing the learning process with part relations helps a dynamic understanding, as illustrated in the first line of~\figref{fig:relation-example} by~\cite{zhao2021graph}. It should be mentioned that most of the prevailing works in fine-grained parsing and even re-identification tend to amplify fixed part regions of the same image, which could be solved by introducing part relationships. 2) Given one image, we usually do not understand which part should network focus on and what are discriminative features for recognition and segmentation. Thus understanding part relationships helps this problem in many ways,~\eg, cross-part relationships within each object helps the understanding of geometric structures, object-part relationship helps the distinguishing of locally similar visual patterns, and part relationships across different images and objects help to learn the semantic consensus. Besides, considering the comparison of visually similar images, these different part-level relationships help the dynamic enhancement of feature extractions. 3) For the semantic ambiguities of fine-grained definition, here we advocate the learning with hierarchical structures, which are still less explored in deep learning works. For example, the human body can be decomposed into heads and bodies, while the head regions can be further subdivided into faces, eyes, and other organs. Building hierarchical trees or graphs helps the holistic geometric structures be more reasonable and is also beneficial for handling the heterogeneous semantic definitions as mentioned in~\figref{fig:challenge}.

\textbf{Relations to fine-grained tasks.} What roles does the \textit{part relationship learning} play in different fine-grained understanding tasks? Here we present a schematic in~\figref{fig:relations} with fine-grained part segmentation, fine-grained recognition and part relationship representations. Part segmentation is the subset of the part relationship representations, while the latter consists of other structural parsing and detection sub-tasks. The intersections of part segmentation and fine-grained recognition are also illustrated in~\secref{sec:part-guidedfgvc}. Methods of this category tend to utilize the part priors to guide the feature extraction, including the fine-grained classification~\cite{zhang2014part,wang2015multiple,huang2016part,he2017weakly,wei2018mask,huang2020interpretable} and re-identifications~\cite{he2019part,kalayeh2018human,meng2020parsing,zhao2021heterogeneous,chen2022pedestrian}. Understanding part relationships without segmenting them as explicit masks or boxes also considerably facilitates the distinguishing of subtle differences, considering the joint region of part relationship representations and fine-grained recognition. Methods using this idea usually adopt the attention mechanisms~\cite{simon2015neural,gonzalez2018semantic,recasens2018learning,fu2017look,ge2019weakly,
wang2020weakly,lam2017fine,he2017weakly,zhao2021part} or graph-based structures~\cite{ji2020attention,nauta2021neural} to guide the learning process. Besides, the other line of works~\cite{
wang2018learning,dubey2018pairwise,aodha19presence,dubey2018maximum,lin2015bilinear,gao2016compact,
li2017factorized,wei2018grassmann,yu2018hierarchical,zheng2019looking,gao2020channel,zhao2021graph} does not rely on part relationships and focuses on the feature representation learning process as mentioned above. It should be noted here we extend the concept of part relationship learning, by incorporating learning with implicit and explicit part localizations, and those methods build explicit part-level relationships. Moreover, fine-grained part segmentation is one of the explicit ways to understand the part relationship procedure, and learning in hierarchical manners or tree/graph structures also leads in promising directions.

\section{Future Directions}\label{sec:direction}
Despite the significant progress made by existing works, there are still many unsolved problems in fine-grained recognition. Here we propose several promising future directions for discussion.

\textbf{Dynamic part relationship learning.} As discussed in~\secref{sec:part-relationsolution}, prevailing part relationship learning works focus on building \textit{static} connections and responses. For example, in fine-grained recognition tasks, networks tend to amplify fixed regions of the same image. Although this helps the discriminative embedding in training data distributions, it faces great challenges when compared to unknown novel testing examples. Besides, the dynamic relationships help the understanding of semantic parsing when objects face occlusions or abnormal gestures.

\textbf{Few-shot fine-grained learning.} The learning of fine-grained classification is based on sufficient training data, while few-shot learning~\cite{lake2015human,fei2006one,miller2000learning} is nowadays an attractive trend for understanding novel concepts with only a few annotated images. As for the few-shot fine-grained classification, there are several advances~\cite{tang2020revisiting,koniusz2021power,huang2020low,zhu2020multi,li2019zero} in this field. However, these works usually follow the N-way K-shot trend, and N is usually set as 5 for the number of categories, indicating the huge gap compared to popular datasets with 500 to 5,000 categories. Besides, few-shot learning also can be regarded as a long-tailed setting where several categories are labeled with sufficient images but a few categories with limited annotations. This few-shot/long-tailed setting is a more realistic challenge that can be taken as a promising direction.

\textbf{Hierarchical structures of fine-grained learning.} In~\figref{fig:challenge} and~\secref{sec:problem-finegrained}, it is noted that the definition of fine-grained settings still exists ambiguous and some subcategories can be further divided into finer levels. Thus to solve this imbalance in semantic space and visual space, we argue for developing hierarchical structures to gradually subdivide these components into meaningful leaf nodes. These leaf nodes can be presented using basic units including pixels and line segments but also basic structures,~\ie, organs of human beings. In other words, the hierarchical structures help to maintain similar concepts in semantic space aligned with that of visual spaces. Several pioneer works~\cite{Zhang_2016_CVPR,xie2015hyper,chang2021your} have explored the tree structures or hyper classes for fine-grained semantic structures. However, how to unify the semantic language embedding with the image-level visual features is still an under-explored problem. One promising direction is to unify the language and visual spaces using contrastive learning and mask modeling, including CLIP~\cite{radford2021learning}, GLIP~\cite{li2022grounded}, and other multi-modality learning methods.

\textbf{3D-aware fine-grained learning.} In addition to the aforementioned 2D-based learning mechanisms, the other promising direction is 3D-aware fine-grained learning. In the semantic parsing of 3D models, many research efforts~\cite{kalogerakis20173d,yu2019partnet,mo2019partnet,qi2017pointnet} have been proposed to parse 3D objects with point cloud, mesh and voxel representations. Thus an interesting question arises here: what is the relationship between 3D parsing models the real-world 2D images? Earlier works collected in~\cite{kittler20053d} propose to use 3D models to aid the recognition of human faces by hand-crafted filters or template learning techniques. In the era of deep learning, several works~\cite{joung2021learning} propose to embed the 3D canonical model with learnable warping parameters to represent diverse 2D images. The emerging field can be further boosted with learnable mechanisms including the Neural rendering field~\cite{mildenhall2021nerf}.

\section{Conclusions}\label{sec:conclusion}
In this paper, we present a comprehensive survey of fine-grained visual parsing tasks from the novel perspective of part relationship learning. In this view, we delve into the connections of two representative fine-grained tasks,~\ie, fine-grained recognition and part segmentation, and propose a new taxonomy to reorganize recent research advances including the conventional methods and deep learning methods. By consolidating these works and popular benchmarks, we propose the universal challenges left in fine-grained visual parsing and make an attempted solution from the view of part relationship learning. Besides, we also point out several promising research directions that can be further explored. We hope these contributions will provide new inspiration to inform future research in the field of fine-grained visual parsing.



\section*{Acknowledgment}
This work was supported in part by the Key-Area Research and Development Program of Guangdong Province under Contract 2021B0101400002, the National Natural Science Foundation of China under contracts No. 62132002, No. 61825101, No. 62202010 and and also supported by the China Postdoctoral Science Foundation No. 2022M710212.

\bibliographystyle{IEEEtran}
\bibliography{survey}

\begin{thebibliography}{100}
\providecommand{\url}[1]{#1}
\csname url@samestyle\endcsname
\providecommand{\newblock}{\relax}
\providecommand{\bibinfo}[2]{#2}
\providecommand{\BIBentrySTDinterwordspacing}{\spaceskip=0pt\relax}
\providecommand{\BIBentryALTinterwordstretchfactor}{4}
\providecommand{\BIBentryALTinterwordspacing}{\spaceskip=\fontdimen2\font plus
\BIBentryALTinterwordstretchfactor\fontdimen3\font minus
  \fontdimen4\font\relax}
\providecommand{\BIBforeignlanguage}[2]{{%
\expandafter\ifx\csname l@#1\endcsname\relax
\typeout{** WARNING: IEEEtran.bst: No hyphenation pattern has been}%
\typeout{** loaded for the language `#1'. Using the pattern for}%
\typeout{** the default language instead.}%
\else
\language=\csname l@#1\endcsname
\fi
#2}}
\providecommand{\BIBdecl}{\relax}
\BIBdecl

\bibitem{deng2009imagenet}
J.~Deng, W.~Dong, R.~Socher, L.-J. Li, K.~Li, and L.~Fei-Fei, ``Imagenet: A
  large-scale hierarchical image database,'' in \emph{IEEE Conference on
  Computer Vision and Pattern Recognition (CVPR)}, 2009, pp. 248--255.

\bibitem{liang2016semantic1}
X.~Liang, X.~Shen, D.~Xiang, J.~Feng, L.~Lin, and S.~Yan, ``Semantic object
  parsing with local-global long short-term memory,'' in \emph{IEEE Conference
  on Computer Vision and Pattern Recognition (CVPR)}, 2016, pp. 3185--3193.

\bibitem{ruan2019devil}
T.~Ruan, T.~Liu, Z.~Huang, Y.~Wei, S.~Wei, and Y.~Zhao, ``Devil in the details:
  Towards accurate single and multiple human parsing,'' in \emph{AAAI
  Conference on Artificial Intelligence (AAAI)}, 2019, pp. 4814--4821.

\bibitem{zhao2021ordinal}
Y.~Zhao, J.~Li, Y.~Zhang, Y.~Song, and Y.~Tian, ``Ordinal multi-task part
  segmentation with recurrent prior generation,'' \emph{IEEE transactions on
  pattern analysis and machine intelligence}, vol.~43, no.~5, pp. 1636--1648,
  2021.

\bibitem{xia2017joint}
F.~Xia, P.~Wang, X.~Chen, and A.~L. Yuille, ``Joint multi-person pose
  estimation and semantic part segmentation,'' in \emph{IEEE Conference on
  Computer Vision and Pattern Recognition (CVPR)}, 2017, pp. 6769--6778.

\bibitem{zhang2014part}
N.~Zhang, J.~Donahue, R.~Girshick, and T.~Darrell, ``Part-based r-cnns for
  fine-grained category detection,'' in \emph{European Conference on Computer
  Vision (ECCV)}.\hskip 1em plus 0.5em minus 0.4em\relax Springer, 2014, pp.
  834--849.

\bibitem{huang2016part}
S.~Huang, Z.~Xu, D.~Tao, and Y.~Zhang, ``Part-stacked cnn for fine-grained
  visual categorization,'' in \emph{IEEE Conference on Computer Vision and
  Pattern Recognition (CVPR)}, 2016, pp. 1173--1182.

\bibitem{he2017weakly}
X.~He and Y.~Peng, ``Weakly supervised learning of part selection model with
  spatial constraints for fine-grained image classification,'' in \emph{AAAI
  Conference on Artificial Intelligence (AAAI)}, 2017.

\bibitem{wei2018mask}
X.-S. Wei, C.-W. Xie, J.~Wu, and C.~Shen, ``Mask-cnn: Localizing parts and
  selecting descriptors for fine-grained bird species categorization,''
  \emph{Pattern Recognition}, vol.~76, pp. 704--714, 2018.

\bibitem{huang2020interpretable}
Z.~Huang and Y.~Li, ``Interpretable and accurate fine-grained recognition via
  region grouping,'' in \emph{IEEE Conference on Computer Vision and Pattern
  Recognition (CVPR)}, 2020, pp. 8662--8672.

\bibitem{bickel2020impacts}
V.~T. Bickel, J.~Aaron, A.~Manconi, S.~Loew, and U.~Mall, ``Impacts drive lunar
  rockfalls over billions of years,'' \emph{Nature communications}, vol.~11,
  no.~1, pp. 1--7, 2020.

\bibitem{sun2022fair1m}
X.~Sun, P.~Wang, Z.~Yan, F.~Xu, R.~Wang, W.~Diao, J.~Chen, J.~Li, Y.~Feng,
  T.~Xu \emph{et~al.}, ``Fair1m: A benchmark dataset for fine-grained object
  recognition in high-resolution remote sensing imagery,'' \emph{ISPRS Journal
  of Photogrammetry and Remote Sensing}, vol. 184, pp. 116--130, 2022.

\bibitem{pakhomov2019deep}
D.~Pakhomov, V.~Premachandran, M.~Allan, M.~Azizian, and N.~Navab, ``Deep
  residual learning for instrument segmentation in robotic surgery,'' in
  \emph{International Workshop on Machine Learning in Medical Imaging}.\hskip
  1em plus 0.5em minus 0.4em\relax Springer, 2019, pp. 566--573.

\bibitem{zhu2008unsupervised}
L.~L. Zhu, C.~Lin, H.~Huang, Y.~Chen, and A.~Yuille, ``Unsupervised structure
  learning: Hierarchical recursive composition, suspicious coincidence and
  competitive exclusion,'' in \emph{European Conference on Computer Vision
  (ECCV)}, 2008, pp. 759--773.

\bibitem{hsieh2010segmentation}
J.-W. Hsieh, C.-H. Chuang, S.-Y. Chen, C.-C. Chen, and K.-C. Fan,
  ``Segmentation of human body parts using deformable triangulation,''
  \emph{IEEE Transactions on Systems, Man, and Cybernetics-Part A: Systems and
  Humans}, vol.~40, no.~3, pp. 596--610, 2010.

\bibitem{wang2011learning}
Y.~Wang, D.~Tran, and Z.~Liao, ``Learning hierarchical poselets for human
  parsing,'' in \emph{IEEE Conference on Computer Vision and Pattern
  Recognition (CVPR)}.\hskip 1em plus 0.5em minus 0.4em\relax IEEE, 2011, pp.
  1705--1712.

\bibitem{car2014parsing}
A.~Y. Wenhao~Lu, Xiaochen~Lian, ``Parsing semantic parts of cars using
  graphical models and segment appearance consistency,'' in \emph{British
  Machine Vision Conference (BMVC)}, 2014.

\bibitem{zhang2012pose}
N.~Zhang, R.~Farrell, and T.~Darrell, ``Pose pooling kernels for sub-category
  recognition,'' in \emph{IEEE Conference on Computer Vision and Pattern
  Recognition (CVPR)}.\hskip 1em plus 0.5em minus 0.4em\relax IEEE, 2012, pp.
  3665--3672.

\bibitem{branson2011strong}
S.~Branson, P.~Perona, and S.~Belongie, ``Strong supervision from weak
  annotation: Interactive training of deformable part models,'' in \emph{IEEE
  International Conference on Computer Vision (ICCV)}.\hskip 1em plus 0.5em
  minus 0.4em\relax IEEE, 2011, pp. 1832--1839.

\bibitem{zhao2017survey}
B.~Zhao, J.~Feng, X.~Wu, and S.~Yan, ``A survey on deep learning-based
  fine-grained object classification and semantic segmentation,''
  \emph{International Journal of Automation and Computing}, vol.~14, no.~2, pp.
  119--135, 2017.

\bibitem{wei2021fine}
X.-S. Wei, Y.-Z. Song, O.~Mac~Aodha, J.~Wu, Y.~Peng, J.~Tang, J.~Yang, and
  S.~Belongie, ``Fine-grained image analysis with deep learning: A survey,''
  \emph{IEEE Transactions on Pattern Analysis and Machine Intelligence}, 2021.

\bibitem{de2021part}
D.~de~Geus, P.~Meletis, C.~Lu, X.~Wen, and G.~Dubbelman, ``Part-aware panoptic
  segmentation,'' in \emph{IEEE Conference on Computer Vision and Pattern
  Recognition (CVPR)}, 2021, pp. 5485--5494.

\bibitem{yamaguchi2012parsing}
K.~Yamaguchi, M.~H. Kiapour, L.~E. Ortiz, and T.~L. Berg, ``Parsing clothing in
  fashion photographs,'' in \emph{IEEE Conference on Computer Vision and
  Pattern Recognition (CVPR)}, 2012, pp. 3570--3577.

\bibitem{chen2014detect}
X.~Chen, R.~Mottaghi, X.~Liu, S.~Fidler, R.~Urtasun, and A.~Yuille, ``Detect
  what you can: Detecting and representing objects using holistic models and
  body parts,'' in \emph{IEEE Conference on Computer Vision and Pattern
  Recognition (CVPR)}, 2014, pp. 1971--1978.

\bibitem{wang2015semantic}
J.~Wang and A.~L. Yuille, ``Semantic part segmentation using compositional
  model combining shape and appearance,'' in \emph{IEEE Conference on Computer
  Vision and Pattern Recognition (CVPR)}, 2015, pp. 1788--1797.

\bibitem{liang2015human}
X.~Liang, C.~Xu, X.~Shen, J.~Yang, S.~Liu, J.~Tang, L.~Lin, and S.~Yan, ``Human
  parsing with contextualized convolutional neural network,'' in \emph{IEEE
  International Conference on Computer Vision (ICCV)}, 2015, pp. 1386--1394.

\bibitem{chen2016attention}
L.-C. Chen, Y.~Yang, J.~Wang, W.~Xu, and A.~L. Yuille, ``Attention to scale:
  Scale-aware semantic image segmentation,'' in \emph{IEEE Conference on
  Computer Vision and Pattern Recognition (CVPR)}, 2016, pp. 3640--3649.

\bibitem{li2017multiple}
J.~Li, J.~Zhao, Y.~Wei, C.~Lang, Y.~Li, T.~Sim, S.~Yan, and J.~Feng,
  ``Multiple-human parsing in the wild,'' \emph{arXiv preprint
  arXiv:1705.07206}, 2017.

\bibitem{liang2018look}
X.~Liang, K.~Gong, X.~Shen, and L.~Lin, ``Look into person: Joint body parsing
  \& pose estimation network and a new benchmark,'' \emph{IEEE Transactions on
  Pattern Analysis and Machine Intelligence}, 2018.

\bibitem{zhou2018adaptive}
Q.~Zhou, X.~Liang, K.~Gong, and L.~Lin, ``Adaptive temporal encoding network
  for video instance-level human parsing,'' in \emph{ACM International
  Conference on Multimedia}, 2018, pp. 1527--1535.

\bibitem{gong2018instance}
K.~Gong, X.~Liang, Y.~Li, Y.~Chen, M.~Yang, and L.~Lin, ``Instance-level human
  parsing via part grouping network,'' in \emph{European Conference on Computer
  Vision (ECCV)}, 2018, pp. 770--785.

\bibitem{zhao2019multi}
Y.~Zhao, J.~Li, Y.~Zhang, and Y.~Tian, ``Multi-class part parsing with joint
  boundary-semantic awareness,'' in \emph{IEEE International Conference on
  Computer Vision (ICCV)}, 2019.

\bibitem{liu2022learning}
Q.~Liu, A.~Kortylewski, Z.~Zhang, Z.~Li, M.~Guo, Q.~Liu, X.~Yuan, J.~Mu,
  W.~Qiu, and A.~Yuille, ``Learning part segmentation through unsupervised
  domain adaptation from synthetic vehicles,'' in \emph{IEEE Conference on
  Computer Vision and Pattern Recognition (CVPR)}, 2022.

\bibitem{michieli2022edge}
U.~Michieli and P.~Zanuttigh, ``Edge-aware graph matching network for
  part-based semantic segmentation,'' \emph{International Journal of Computer
  Vision}, vol. 130, no.~11, pp. 2797--2821, 2022.

\bibitem{zhou2017scene}
B.~Zhou, H.~Zhao, X.~Puig, S.~Fidler, A.~Barriuso, and A.~Torralba, ``Scene
  parsing through ade20k dataset,'' in \emph{IEEE Conference on Computer Vision
  and Pattern Recognition (CVPR)}, 2017, pp. 633--641.

\bibitem{fang2018weakly}
H.-S. Fang, G.~Lu, X.~Fang, J.~Xie, Y.-W. Tai, and C.~Lu, ``Weakly and semi
  supervised human body part parsing via pose-guided knowledge transfer,'' in
  \emph{IEEE Conference on Computer Vision and Pattern Recognition (CVPR)},
  2018, pp. 70--78.

\bibitem{xia2016zoom}
F.~Xia, P.~Wang, L.-C. Chen, and A.~L. Yuille, ``Zoom better to see clearer:
  Human and object parsing with hierarchical auto-zoom net,'' in \emph{European
  Conference on Computer Vision (ECCV)}, 2016, pp. 648--663.

\bibitem{nie2018mutual}
X.~Nie, J.~Feng, and S.~Yan, ``Mutual learning to adapt for joint human parsing
  and pose estimation,'' in \emph{European Conference on Computer Vision
  (ECCV)}, 2018, pp. 502--517.

\bibitem{li2021multi}
J.~Li, J.~Zhao, C.~Lang, Y.~Li, Y.~Wei, G.~Guo, T.~Sim, S.~Yan, and J.~Feng,
  ``Multi-human parsing with a graph-based generative adversarial model,''
  \emph{ACM Transactions on Multimedia Computing, Communications, and
  Applications (TOMM)}, vol.~17, no.~1, pp. 1--21, 2021.

\bibitem{wang2019learning}
W.~Wang, Z.~Zhang, S.~Qi, J.~Shen, Y.~Pang, and L.~Shao, ``Learning
  compositional neural information fusion for human parsing,'' in \emph{IEEE
  International Conference on Computer Vision (ICCV)}, October 2019.

\bibitem{gong2019graphonomy}
K.~Gong, Y.~Gao, X.~Liang, X.~Shen, M.~Wang, and L.~Lin, ``Graphonomy:
  Universal human parsing via graph transfer learning,'' in \emph{IEEE
  Conference on Computer Vision and Pattern Recognition (CVPR)}, 2019, pp.
  7450--7459.

\bibitem{liu2019braidnet}
X.~Liu, M.~Zhang, W.~Liu, J.~Song, and T.~Mei, ``Braidnet: Braiding semantics
  and details for accurate human parsing,'' in \emph{ACM International
  Conference on Multimedia}, 2019, pp. 338--346.

\bibitem{wang2020hierarchical}
W.~Wang, H.~Zhu, J.~Dai, Y.~Pang, J.~Shen, and L.~Shao, ``Hierarchical human
  parsing with typed part-relation reasoning,'' in \emph{IEEE Conference on
  Computer Vision and Pattern Recognition (CVPR)}, June 2020.

\bibitem{zhou2021differentiable}
T.~Zhou, W.~Wang, S.~Liu, Y.~Yang, and L.~Van~Gool, ``Differentiable
  multi-granularity human representation learning for instance-aware human
  semantic parsing,'' in \emph{IEEE Conference on Computer Vision and Pattern
  Recognition (CVPR)}, 2021, pp. 1622--1631.

\bibitem{zeng2021neural}
D.~Zeng, Y.~Huang, Q.~Bao, J.~Zhang, C.~Su, and W.~Liu, ``Neural architecture
  search for joint human parsing and pose estimation,'' in \emph{IEEE
  International Conference on Computer Vision (ICCV)}, 2021, pp.
  11\,385--11\,394.

\bibitem{liu2021hierarchical}
Y.~Liu, S.~Zhang, J.~Yang, and P.~Yuen, ``Hierarchical information passing
  based noise-tolerant hybrid learning for semi-supervised human parsing,'' in
  \emph{AAAI Conference on Artificial Intelligence (AAAI)}, vol.~35, no.~3,
  2021, pp. 2207--2215.

\bibitem{zhao2020fine}
J.~Zhao, J.~Li, H.~Liu, S.~Yan, and J.~Feng, ``Fine-grained multi-human
  parsing,'' \emph{International Journal of Computer Vision}, vol. 128, no.~8,
  pp. 2185--2203, 2020.

\bibitem{yang2019parsing}
L.~Yang, Q.~Song, Z.~Wang, and M.~Jiang, ``Parsing r-cnn for instance-level
  human analysis,'' in \emph{IEEE Conference on Computer Vision and Pattern
  Recognition (CVPR)}, June 2019.

\bibitem{li2020self}
P.~Li, Y.~Xu, Y.~Wei, and Y.~Yang, ``Self-correction for human parsing,''
  \emph{IEEE Transactions on Pattern Analysis and Machine Intelligence}, 2020.

\bibitem{he2020grapy}
H.~He, J.~Zhang, Q.~Zhang, and D.~Tao, ``Grapy-ml: Graph pyramid mutual
  learning for cross-dataset human parsing,'' in \emph{AAAI Conference on
  Artificial Intelligence (AAAI)}, vol.~34, no.~07, 2020, pp. 10\,949--10\,956.

\bibitem{ji2020learning}
R.~Ji, D.~Du, L.~Zhang, L.~Wen, Y.~Wu, C.~Zhao, F.~Huang, and S.~Lyu,
  ``Learning semantic neural tree for human parsing,'' in \emph{European
  Conference on Computer Vision (ECCV)}.\hskip 1em plus 0.5em minus 0.4em\relax
  Springer, 2020, pp. 205--221.

\bibitem{zhang2020human}
S.~Zhang, G.-J. Qi, X.~Cao, Z.~Song, and J.~Zhou, ``Human parsing with
  pyramidical gather-excite context,'' \emph{IEEE Transactions on Circuits and
  Systems for Video Technology}, vol.~31, no.~3, pp. 1016--1030, 2020.

\bibitem{zhang2020part}
X.~Zhang, Y.~Chen, B.~Zhu, J.~Wang, and M.~Tang, ``Part-aware context network
  for human parsing,'' in \emph{IEEE Conference on Computer Vision and Pattern
  Recognition (CVPR)}, June 2020.

\bibitem{loesch2021describe}
A.~Loesch and R.~Audigier, ``Describe me if you can! characterized
  instance-level human parsing,'' in \emph{IEEE Conference on Image Processing
  (ICIP)}.\hskip 1em plus 0.5em minus 0.4em\relax IEEE, 2021, pp. 2528--2532.

\bibitem{song2017embedding}
Y.~Song, X.~Chen, J.~Li, and Q.~Zhao, ``Embedding 3d geometric features for
  rigid object part segmentation,'' in \emph{IEEE International Conference on
  Computer Vision (ICCV)}, 2017, pp. 580--588.

\bibitem{wang2015joint}
P.~Wang, X.~Shen, Z.~Lin, S.~Cohen, B.~Price, and A.~L. Yuille, ``Joint object
  and part segmentation using deep learned potentials,'' in \emph{IEEE
  International Conference on Computer Vision (ICCV)}, 2015, pp. 1573--1581.

\bibitem{naha2021part}
S.~Naha, Q.~Xiao, P.~Banik, M.~A. Reza, and D.~J. Crandall, ``Part segmentation
  of unseen objects using keypoint guidance,'' in \emph{Proceedings of the
  IEEE/CVF Winter Conference on Applications of Computer Vision (WACV)}, 2021,
  pp. 1742--1750.

\bibitem{wu2019keypoint}
Z.~Wu, G.~Lin, and J.~Cai, ``Keypoint based weakly supervised human parsing,''
  \emph{Image and Vision Computing}, vol.~91, p. 103801, 2019.

\bibitem{yang2021weakly}
Z.~Yang, Y.~Li, L.~Yang, N.~Zhang, and J.~Luo, ``Weakly supervised body part
  segmentation with pose based part priors,'' in \emph{2020 25th International
  Conference on Pattern Recognition (ICPR)}.\hskip 1em plus 0.5em minus
  0.4em\relax IEEE, 2021, pp. 286--293.

\bibitem{zhao2022pose}
Y.~Zhao, J.~Li, Y.~Zhang, and Y.~Tian, ``From pose to part: Weakly-supervised
  pose evolution for human part segmentation,'' \emph{IEEE Transactions on
  Pattern Analysis and Machine Intelligence}, 2022.

\bibitem{yang2021learning}
Y.~Yang, X.~Cheng, H.~Bilen, and X.~Ji, ``Learning to annotate part
  segmentation with gradient matching,'' in \emph{International Conference on
  Learning Representations}, 2021.

\bibitem{gonzalez2018semantic}
A.~Gonzalez-Garcia, D.~Modolo, and V.~Ferrari, ``Do semantic parts emerge in
  convolutional neural networks?'' \emph{International Journal of Computer
  Vision}, vol. 126, no.~5, pp. 476--494, 2018.

\bibitem{lorenz2019unsupervised}
D.~Lorenz, L.~Bereska, T.~Milbich, and B.~Ommer, ``Unsupervised part-based
  disentangling of object shape and appearance,'' in \emph{IEEE Conference on
  Computer Vision and Pattern Recognition (CVPR)}, 2019, pp. 10\,955--10\,964.

\bibitem{hung2019scops}
W.-C. Hung, V.~Jampani, S.~Liu, P.~Molchanov, M.-H. Yang, and J.~Kautz,
  ``Scops: Self-supervised co-part segmentation,'' in \emph{IEEE Conference on
  Computer Vision and Pattern Recognition (CVPR)}, 2019, pp. 869--878.

\bibitem{gao2021unsupervised}
Q.~Gao, B.~Wang, L.~Liu, and B.~Chen, ``Unsupervised co-part segmentation
  through assembly,'' in \emph{International Conference on Machine Learning
  (ICML)}.\hskip 1em plus 0.5em minus 0.4em\relax PMLR, 2021, pp. 3576--3586.

\bibitem{liu2021unsupervised}
S.~Liu, L.~Zhang, X.~Yang, H.~Su, and J.~Zhu, ``Unsupervised part segmentation
  through disentangling appearance and shape,'' in \emph{IEEE Conference on
  Computer Vision and Pattern Recognition (CVPR)}, June 2021, pp. 8355--8364.

\bibitem{choudhury2021unsupervised}
S.~Choudhury, I.~Laina, C.~Rupprecht, and A.~Vedaldi, ``Unsupervised part
  discovery from contrastive reconstruction,'' \emph{Advances in Neural
  Information Processing Systems (NeurIPS)}, vol.~34, 2021.

\bibitem{michieli2020gmnet}
U.~Michieli, E.~Borsato, L.~Rossi, and P.~Zanuttigh, ``Gmnet: Graph matching
  network for large scale part semantic segmentation in the wild,'' in
  \emph{European Conference on Computer Vision (ECCV)}, 2020, pp. 397--414.

\bibitem{tan2021confident}
X.~Tan, J.~Xu, Z.~Ye, J.~Hao, and L.~Ma, ``Confident semantic ranking loss for
  part parsing,'' in \emph{2021 IEEE International Conference on Multimedia and
  Expo (ICME)}.\hskip 1em plus 0.5em minus 0.4em\relax IEEE, 2021, pp. 1--6.

\bibitem{singh2022float}
R.~Singh, P.~Gupta, P.~Shenoy, and R.~Sarvadevabhatla, ``Float: Factorized
  learning of object attributes for improved multi-object multi-part scene
  parsing,'' \emph{IEEE Conference on Computer Vision and Pattern Recognition
  (CVPR)}, 2022.

\bibitem{kirillov2019panoptic}
A.~Kirillov, K.~He, R.~Girshick, C.~Rother, and P.~Doll{\'a}r, ``Panoptic
  segmentation,'' in \emph{IEEE Conference on Computer Vision and Pattern
  Recognition (CVPR)}, 2019, pp. 9404--9413.

\bibitem{Everingham15}
M.~Everingham, S.~M.~A. Eslami, L.~Van~Gool, C.~K.~I. Williams, J.~Winn, and
  A.~Zisserman, ``The pascal visual object classes challenge: A
  retrospective,'' in \emph{IEEE International Conference on Computer Vision
  (ICCV)}, vol. 111, no.~1, 2015, pp. 98--136.

\bibitem{Cordts2016Cityscapes}
M.~Cordts, M.~Omran, S.~Ramos, T.~Rehfeld, M.~Enzweiler, R.~Benenson,
  U.~Franke, S.~Roth, and B.~Schiele, ``The cityscapes dataset for semantic
  urban scene understanding,'' in \emph{IEEE Conference on Computer Vision and
  Pattern Recognition (CVPR)}, 2016.

\bibitem{felzenszwalb2010object}
P.~F. Felzenszwalb, R.~B. Girshick, D.~McAllester, and D.~Ramanan, ``Object
  detection with discriminatively trained part-based models,'' \emph{IEEE
  Transactions on Pattern Analysis and Machine Intelligence}, vol.~32, no.~9,
  pp. 1627--1645, 2010.

\bibitem{eslami2012generative}
S.~Eslami and C.~Williams, ``A generative model for parts-based object
  segmentation,'' in \emph{Advances in Neural Information Processing Systems
  (NeurIPS)}, 2012, pp. 100--107.

\bibitem{liu2014fashion}
S.~Liu, J.~Feng, C.~Domokos, H.~Xu, J.~Huang, Z.~Hu, and S.~Yan, ``Fashion
  parsing with weak color-category labels,'' \emph{IEEE Transactions on
  Multimedia}, vol.~16, no.~1, pp. 253--265, 2014.

\bibitem{meng2017seeds}
F.~Meng, H.~Li, Q.~Wu, K.~N. Ngan, and J.~Cai, ``Seeds-based part segmentation
  by seeds propagation and region convexity decomposition,'' \emph{IEEE
  Transactions on Multimedia}, vol.~20, no.~2, pp. 310--322, 2017.

\bibitem{desai2012detecting}
C.~Desai and D.~Ramanan, ``Detecting actions, poses, and objects with
  relational phraselets,'' in \emph{European Conference on Computer Vision
  (ECCV)}.\hskip 1em plus 0.5em minus 0.4em\relax Springer, 2012, pp. 158--172.

\bibitem{azizpour2012object}
H.~Azizpour and I.~Laptev, ``Object detection using strongly-supervised
  deformable part models,'' in \emph{European Conference on Computer Vision
  (ECCV)}, 2012, pp. 836--849.

\bibitem{dong2015parsing}
J.~Dong, Q.~Chen, Z.~Huang, J.~Yang, and S.~Yan, ``Parsing based on parselets:
  A unified deformable mixture model for human parsing,'' \emph{IEEE
  Transactions on Pattern Analysis and Machine Intelligence}, vol.~38, no.~1,
  pp. 88--101, 2015.

\bibitem{xia2016pose}
F.~Xia, J.~Zhu, P.~Wang, and A.~Yuille, ``Pose-guided human parsing by an
  and/or graph using pose-context features,'' in \emph{AAAI Conference on
  Artificial Intelligence (AAAI)}, vol.~30, no.~1, 2016.

\bibitem{he2016deep}
K.~He, X.~Zhang, S.~Ren, and J.~Sun, ``Deep residual learning for image
  recognition,'' in \emph{IEEE Conference on Computer Vision and Pattern
  Recognition (CVPR)}, 2016, pp. 770--778.

\bibitem{simonyan2014very}
K.~Simonyan and A.~Zisserman, ``Very deep convolutional networks for
  large-scale image recognition,'' \emph{arXiv preprint arXiv:1409.1556}, 2014.

\bibitem{he2015delving}
K.~He, X.~Zhang, S.~Ren, and J.~Sun, ``Delving deep into rectifiers: Surpassing
  human-level performance on imagenet classification,'' in \emph{IEEE
  International Conference on Computer Vision (ICCV)}, 2015, pp. 1026--1034.

\bibitem{long2015fully}
J.~Long, E.~Shelhamer, and T.~Darrell, ``Fully convolutional networks for
  semantic segmentation,'' in \emph{IEEE Conference on Computer Vision and
  Pattern Recognition (CVPR)}, 2015, pp. 3431--3440.

\bibitem{yang2011articulated}
Y.~Yang and D.~Ramanan, ``Articulated pose estimation with flexible
  mixtures-of-parts,'' in \emph{IEEE Conference on Computer Vision and Pattern
  Recognition (CVPR)}, 2011, pp. 1385--1392.

\bibitem{dong2014towards}
J.~Dong, Q.~Chen, X.~Shen, J.~Yang, and S.~Yan, ``Towards unified human parsing
  and pose estimation,'' in \emph{IEEE Conference on Computer Vision and
  Pattern Recognition (CVPR)}, 2014, pp. 843--850.

\bibitem{chen2018deeplab}
L.-C. Chen, G.~Papandreou, I.~Kokkinos, K.~Murphy, and A.~L. Yuille, ``Deeplab:
  Semantic image segmentation with deep convolutional nets, atrous convolution,
  and fully connected crfs,'' \emph{IEEE Transactions on Pattern Analysis and
  Machine Intelligence}, vol.~40, no.~4, pp. 834--848, 2018.

\bibitem{chen2017rethinking}
L.-C. Chen, G.~Papandreou, F.~Schroff, and H.~Adam, ``Rethinking atrous
  convolution for semantic image segmentation,'' \emph{arXiv:1706.05587}, 2017.

\bibitem{liang2016semantic}
X.~Liang, X.~Shen, J.~Feng, L.~Lin, and S.~Yan, ``Semantic object parsing with
  graph lstm,'' in \emph{European Conference on Computer Vision (ECCV)}.\hskip
  1em plus 0.5em minus 0.4em\relax Springer, 2016, pp. 125--143.

\bibitem{wah2011caltech}
C.~Wah, S.~Branson, P.~Welinder, P.~Perona, and S.~Belongie, ``The caltech-ucsd
  birds-200-2011 dataset,'' 2011.

\bibitem{nilsback2008automated}
M.-E. Nilsback and A.~Zisserman, ``Automated flower classification over a large
  number of classes,'' in \emph{2008 Sixth Indian Conference on Computer
  Vision, Graphics \& Image Processing}.\hskip 1em plus 0.5em minus 0.4em\relax
  IEEE, 2008, pp. 722--729.

\bibitem{khosla2011novel}
A.~Khosla, N.~Jayadevaprakash, B.~Yao, and F.-F. Li, ``Novel dataset for
  fine-grained image categorization: Stanford dogs,'' in \emph{Proc. CVPR
  workshop on fine-grained visual categorization (FGVC)}, vol.~2, no.~1.\hskip
  1em plus 0.5em minus 0.4em\relax Citeseer, 2011.

\bibitem{krause20133d}
J.~Krause, M.~Stark, J.~Deng, and L.~Fei-Fei, ``3d object representations for
  fine-grained categorization,'' in \emph{Proceedings of the IEEE International
  Conference on Computer Vision Workshops}, 2013, pp. 554--561.

\bibitem{maji2013fine}
S.~Maji, E.~Rahtu, J.~Kannala, M.~Blaschko, and A.~Vedaldi, ``Fine-grained
  visual classification of aircraft,'' \emph{arXiv preprint arXiv:1306.5151},
  2013.

\bibitem{bossard2014food}
L.~Bossard, M.~Guillaumin, and L.~V. Gool, ``Food-101--mining discriminative
  components with random forests,'' in \emph{European Conference on Computer
  Vision (ECCV)}.\hskip 1em plus 0.5em minus 0.4em\relax Springer, 2014, pp.
  446--461.

\bibitem{berg2014birdsnap}
T.~Berg, J.~Liu, S.~Woo~Lee, M.~L. Alexander, D.~W. Jacobs, and P.~N.
  Belhumeur, ``Birdsnap: Large-scale fine-grained visual categorization of
  birds,'' in \emph{IEEE Conference on Computer Vision and Pattern Recognition
  (CVPR)}, 2014, pp. 2011--2018.

\bibitem{van2015building}
G.~Van~Horn, S.~Branson, R.~Farrell, S.~Haber, J.~Barry, P.~Ipeirotis,
  P.~Perona, and S.~Belongie, ``Building a bird recognition app and large scale
  dataset with citizen scientists: The fine print in fine-grained dataset
  collection,'' in \emph{IEEE Conference on Computer Vision and Pattern
  Recognition (CVPR)}, 2015, pp. 595--604.

\bibitem{yang2015large}
L.~Yang, P.~Luo, C.~Change~Loy, and X.~Tang, ``A large-scale car dataset for
  fine-grained categorization and verification,'' in \emph{IEEE Conference on
  Computer Vision and Pattern Recognition (CVPR)}, 2015, pp. 3973--3981.

\bibitem{liu2016deepfashion}
Z.~Liu, P.~Luo, S.~Qiu, X.~Wang, and X.~Tang, ``Deepfashion: Powering robust
  clothes recognition and retrieval with rich annotations,'' in \emph{IEEE
  Conference on Computer Vision and Pattern Recognition (CVPR)}, 2016, pp.
  1096--1104.

\bibitem{van2018inaturalist}
G.~Van~Horn, O.~Mac~Aodha, Y.~Song, Y.~Cui, C.~Sun, A.~Shepard, H.~Adam,
  P.~Perona, and S.~Belongie, ``The inaturalist species classification and
  detection dataset,'' in \emph{IEEE Conference on Computer Vision and Pattern
  Recognition (CVPR)}, 2018, pp. 8769--8778.

\bibitem{sun2018multi}
M.~Sun, Y.~Yuan, F.~Zhou, and E.~Ding, ``Multi-attention multi-class constraint
  for fine-grained image recognition,'' in \emph{European Conference on
  Computer Vision (ECCV)}, 2018, pp. 805--821.

\bibitem{van2021benchmarking}
G.~Van~Horn, E.~Cole, S.~Beery, K.~Wilber, S.~Belongie, and O.~Mac~Aodha,
  ``Benchmarking representation learning for natural world image collections,''
  in \emph{IEEE Conference on Computer Vision and Pattern Recognition (CVPR)},
  2021, pp. 12\,884--12\,893.

\bibitem{zhuang2018wildfish}
P.~Zhuang, Y.~Wang, and Y.~Qiao, ``Wildfish: A large benchmark for fish
  recognition in the wild,'' in \emph{ACM International Conference on
  Multimedia}, 2018, pp. 1301--1309.

\bibitem{weyand2020google}
T.~Weyand, A.~Araujo, B.~Cao, and J.~Sim, ``Google landmarks dataset v2-a
  large-scale benchmark for instance-level recognition and retrieval,'' in
  \emph{IEEE Conference on Computer Vision and Pattern Recognition (CVPR)},
  2020, pp. 2575--2584.

\bibitem{yao2011combining}
B.~Yao, A.~Khosla, and L.~Fei-Fei, ``Combining randomization and discrimination
  for fine-grained image categorization,'' in \emph{IEEE Conference on Computer
  Vision and Pattern Recognition (CVPR)}.\hskip 1em plus 0.5em minus
  0.4em\relax IEEE, 2011, pp. 1577--1584.

\bibitem{yao2012codebook}
B.~Yao, G.~Bradski, and L.~Fei-Fei, ``A codebook-free and annotation-free
  approach for fine-grained image categorization,'' in \emph{IEEE Conference on
  Computer Vision and Pattern Recognition (CVPR)}.\hskip 1em plus 0.5em minus
  0.4em\relax IEEE, 2012, pp. 3466--3473.

\bibitem{goring2014nonparametric}
C.~Goring, E.~Rodner, A.~Freytag, and J.~Denzler, ``Nonparametric part transfer
  for fine-grained recognition,'' in \emph{IEEE Conference on Computer Vision
  and Pattern Recognition (CVPR)}, 2014, pp. 2489--2496.

\bibitem{wah2011multiclass}
C.~Wah, S.~Branson, P.~Perona, and S.~Belongie, ``Multiclass recognition and
  part localization with humans in the loop,'' in \emph{IEEE International
  Conference on Computer Vision (ICCV)}.\hskip 1em plus 0.5em minus 0.4em\relax
  IEEE, 2011, pp. 2524--2531.

\bibitem{he2019part}
B.~He, J.~Li, Y.~Zhao, and Y.~Tian, ``Part-regularized near-duplicate vehicle
  re-identification,'' in \emph{IEEE Conference on Computer Vision and Pattern
  Recognition (CVPR)}, 2019, pp. 3997--4005.

\bibitem{peng2017object}
Y.~Peng, X.~He, and J.~Zhao, ``Object-part attention model for fine-grained
  image classification,'' \emph{IEEE Transactions on Image Processing},
  vol.~27, no.~3, pp. 1487--1500, 2017.

\bibitem{wang2015multiple}
D.~Wang, Z.~Shen, J.~Shao, W.~Zhang, X.~Xue, and Z.~Zhang, ``Multiple
  granularity descriptors for fine-grained categorization,'' in
  \emph{Proceedings of the IEEE international conference on computer vision},
  2015, pp. 2399--2406.

\bibitem{krause2015fine}
J.~Krause, H.~Jin, J.~Yang, and L.~Fei-Fei, ``Fine-grained recognition without
  part annotations,'' in \emph{IEEE Conference on Computer Vision and Pattern
  Recognition (CVPR)}, 2015, pp. 5546--5555.

\bibitem{zhao2021graph}
Y.~Zhao, K.~Yan, F.~Huang, and J.~Li, ``Graph-based high-order relation
  discovery for fine-grained recognition,'' in \emph{Proceedings of the
  IEEE/CVF Conference on Computer Vision and Pattern Recognition}, 2021, pp.
  15\,079--15\,088.

\bibitem{lin2015bilinear}
T.-Y. Lin, A.~RoyChowdhury, and S.~Maji, ``Bilinear cnn models for fine-grained
  visual recognition,'' in \emph{IEEE International Conference on Computer
  Vision (ICCV)}, 2015, pp. 1449--1457.

\bibitem{gao2016compact}
Y.~Gao, O.~Beijbom, N.~Zhang, and T.~Darrell, ``Compact bilinear pooling,'' in
  \emph{IEEE Conference on Computer Vision and Pattern Recognition (CVPR)},
  2016, pp. 317--326.

\bibitem{li2017factorized}
Y.~Li, N.~Wang, J.~Liu, and X.~Hou, ``Factorized bilinear models for image
  recognition,'' in \emph{IEEE International Conference on Computer Vision
  (ICCV)}, 2017, pp. 2079--2087.

\bibitem{wei2018grassmann}
X.~Wei, Y.~Zhang, Y.~Gong, J.~Zhang, and N.~Zheng, ``Grassmann pooling as
  compact homogeneous bilinear pooling for fine-grained visual
  classification,'' in \emph{European Conference on Computer Vision (ECCV)},
  2018, pp. 355--370.

\bibitem{zheng2019looking}
H.~Zheng, J.~Fu, Z.-J. Zha, and J.~Luo, ``Looking for the devil in the details:
  Learning trilinear attention sampling network for fine-grained image
  recognition,'' in \emph{IEEE Conference on Computer Vision and Pattern
  Recognition (CVPR)}, 2019, pp. 5012--5021.

\bibitem{gao2020channel}
Y.~Gao, X.~Han, X.~Wang, W.~Huang, and M.~Scott, ``Channel interaction networks
  for fine-grained image categorization,'' in \emph{AAAI Conference on
  Artificial Intelligence (AAAI)}, 2020, pp. 10\,818--10\,825.

\bibitem{wang2018non}
X.~Wang, R.~Girshick, A.~Gupta, and K.~He, ``Non-local neural networks,'' in
  \emph{Proceedings of the IEEE conference on computer vision and pattern
  recognition}, 2018, pp. 7794--7803.

\bibitem{simon2015neural}
M.~Simon and E.~Rodner, ``Neural activation constellations: Unsupervised part
  model discovery with convolutional networks,'' in \emph{IEEE International
  Conference on Computer Vision (ICCV)}, 2015, pp. 1143--1151.

\bibitem{zhang2016weakly}
Y.~Zhang, X.-S. Wei, J.~Wu, J.~Cai, J.~Lu, V.-A. Nguyen, and M.~N. Do, ``Weakly
  supervised fine-grained categorization with part-based image
  representation,'' \emph{IEEE Transactions on Image Processing}, vol.~25,
  no.~4, pp. 1713--1725, 2016.

\bibitem{fu2017look}
J.~Fu, H.~Zheng, and T.~Mei, ``Look closer to see better: Recurrent attention
  convolutional neural network for fine-grained image recognition,'' in
  \emph{IEEE Conference on Computer Vision and Pattern Recognition (CVPR)},
  2017, pp. 4438--4446.

\bibitem{recasens2018learning}
A.~Recasens, P.~Kellnhofer, S.~Stent, W.~Matusik, and A.~Torralba, ``Learning
  to zoom: a saliency-based sampling layer for neural networks,'' in
  \emph{European Conference on Computer Vision (ECCV)}, 2018, pp. 51--66.

\bibitem{wang2020weakly}
Z.~Wang, S.~Wang, S.~Yang, H.~Li, J.~Li, and Z.~Li, ``Weakly supervised
  fine-grained image classification via guassian mixture model oriented
  discriminative learning,'' in \emph{IEEE Conference on Computer Vision and
  Pattern Recognition (CVPR)}, 2020, pp. 9749--9758.

\bibitem{ge2019weakly}
W.~Ge, X.~Lin, and Y.~Yu, ``Weakly supervised complementary parts models for
  fine-grained image classification from the bottom up,'' in \emph{IEEE
  Conference on Computer Vision and Pattern Recognition (CVPR)}, 2019, pp.
  3034--3043.

\bibitem{sun2020fine}
G.~Sun, H.~Cholakkal, S.~Khan, F.~Khan, and L.~Shao, ``Fine-grained
  recognition: Accounting for subtle differences between similar classes,'' in
  \emph{AAAI Conference on Artificial Intelligence (AAAI)}, vol.~34, no.~07,
  2020, pp. 12\,047--12\,054.

\bibitem{zheng2019learning}
H.~Zheng, J.~Fu, Z.-J. Zha, J.~Luo, and T.~Mei, ``Learning rich part
  hierarchies with progressive attention networks for fine-grained image
  recognition,'' \emph{IEEE Transactions on Image Processing}, vol.~29, pp.
  476--488, 2019.

\bibitem{ding2019selective}
Y.~Ding, Y.~Zhou, Y.~Zhu, Q.~Ye, and J.~Jiao, ``Selective sparse sampling for
  fine-grained image recognition,'' in \emph{IEEE International Conference on
  Computer Vision (ICCV)}, 2019, pp. 6599--6608.

\bibitem{wang2020graph}
Z.~Wang, S.~Wang, H.~Li, Z.~Dou, and J.~Li, ``Graph-propagation based
  correlation learning for weakly supervised fine-grained image
  classification,'' in \emph{AAAI Conference on Artificial Intelligence
  (AAAI)}, vol.~34, no.~07, 2020, pp. 12\,289--12\,296.

\bibitem{lam2017fine}
M.~Lam, B.~Mahasseni, and S.~Todorovic, ``Fine-grained recognition as hsnet
  search for informative image parts,'' in \emph{IEEE Conference on Computer
  Vision and Pattern Recognition (CVPR)}, 2017, pp. 2520--2529.

\bibitem{zhao2021part}
Y.~Zhao, J.~Li, X.~Chen, and Y.~Tian, ``Part-guided relational transformers for
  fine-grained visual recognition,'' \emph{IEEE Transactions on Image
  Processing}, vol.~30, pp. 9470--9481, 2021.

\bibitem{ji2020attention}
R.~Ji, L.~Wen, L.~Zhang, D.~Du, Y.~Wu, C.~Zhao, X.~Liu, and F.~Huang,
  ``Attention convolutional binary neural tree for fine-grained visual
  categorization,'' in \emph{IEEE Conference on Computer Vision and Pattern
  Recognition (CVPR)}, 2020, pp. 10\,468--10\,477.

\bibitem{nauta2021neural}
M.~Nauta, R.~van Bree, and C.~Seifert, ``Neural prototype trees for
  interpretable fine-grained image recognition,'' in \emph{Proceedings of the
  IEEE/CVF Conference on Computer Vision and Pattern Recognition}, 2021, pp.
  14\,933--14\,943.

\bibitem{yang2018learning}
Z.~Yang, T.~Luo, D.~Wang, Z.~Hu, J.~Gao, and L.~Wang, ``Learning to navigate
  for fine-grained classification,'' in \emph{European Conference on Computer
  Vision (ECCV)}, 2018, pp. 420--435.

\bibitem{wang2018learning}
Y.~Wang, V.~I. Morariu, and L.~S. Davis, ``Learning a discriminative filter
  bank within a cnn for fine-grained recognition,'' in \emph{IEEE Conference on
  Computer Vision and Pattern Recognition (CVPR)}, 2018, pp. 4148--4157.

\bibitem{dubey2018pairwise}
A.~Dubey, O.~Gupta, P.~Guo, R.~Raskar, R.~Farrell, and N.~Naik, ``Pairwise
  confusion for fine-grained visual classification,'' in \emph{European
  Conference on Computer Vision (ECCV)}, 2018, pp. 70--86.

\bibitem{aodha19presence}
O.~M. Aodha, E.~Cole, and P.~Perona, ``Presence-only geographical priors for
  fine-grained image classification,'' in \emph{IEEE International Conference
  on Computer Vision (ICCV)}, 2019, pp. 9595--9605.

\bibitem{dubey2018maximum}
A.~Dubey, O.~Gupta, R.~Raskar, and N.~Naik, ``Maximum-entropy fine grained
  classification,'' in \emph{Advances in Neural Information Processing Systems
  (NeurIPS)}, 2018, pp. 637--647.

\bibitem{cui2018large}
Y.~Cui, Y.~Song, C.~Sun, A.~Howard, and S.~Belongie, ``Large scale fine-grained
  categorization and domain-specific transfer learning,'' in \emph{IEEE
  Conference on Computer Vision and Pattern Recognition (CVPR)}, 2018, pp.
  4109--4118.

\bibitem{zheng2019towards}
X.~Zheng, R.~Ji, X.~Sun, B.~Zhang, Y.~Wu, and F.~Huang, ``Towards optimal fine
  grained retrieval via decorrelated centralized loss with normalize-scale
  layer,'' in \emph{AAAI Conference on Artificial Intelligence (AAAI)},
  vol.~33, no.~01, 2019, pp. 9291--9298.

\bibitem{wei2017selective}
X.-S. Wei, J.-H. Luo, J.~Wu, and Z.-H. Zhou, ``Selective convolutional
  descriptor aggregation for fine-grained image retrieval,'' \emph{IEEE
  Transactions on Image Processing}, vol.~26, no.~6, pp. 2868--2881, 2017.

\bibitem{yu2018hierarchical}
C.~Yu, X.~Zhao, Q.~Zheng, P.~Zhang, and X.~You, ``Hierarchical bilinear pooling
  for fine-grained visual recognition,'' in \emph{European Conference on
  Computer Vision (ECCV)}, 2018, pp. 574--589.

\bibitem{zhang2019learning}
L.~Zhang, S.~Huang, W.~Liu, and D.~Tao, ``Learning a mixture of
  granularity-specific experts for fine-grained categorization,'' in \emph{IEEE
  International Conference on Computer Vision (ICCV)}, 2019, pp. 8331--8340.

\bibitem{kong2017low}
S.~Kong and C.~Fowlkes, ``Low-rank bilinear pooling for fine-grained
  classification,'' in \emph{IEEE Conference on Computer Vision and Pattern
  Recognition (CVPR)}, 2017, pp. 365--374.

\bibitem{chen2019destruction}
Y.~Chen, Y.~Bai, W.~Zhang, and T.~Mei, ``Destruction and construction learning
  for fine-grained image recognition,'' in \emph{IEEE Conference on Computer
  Vision and Pattern Recognition (CVPR)}, 2019, pp. 5157--5166.

\bibitem{luo2019cross}
W.~Luo, X.~Yang, X.~Mo, Y.~Lu, L.~S. Davis, J.~Li, J.~Yang, and S.-N. Lim,
  ``Cross-x learning for fine-grained visual categorization,'' in \emph{IEEE
  International Conference on Computer Vision (ICCV)}, 2019, pp. 8242--8251.

\bibitem{zhuang2020learning}
P.~Zhuang, Y.~Wang, and Y.~Qiao, ``Learning attentive pairwise interaction for
  fine-grained classification,'' in \emph{AAAI Conference on Artificial
  Intelligence (AAAI)}, vol.~34, no.~07, 2020, pp. 13\,130--13\,137.

\bibitem{liu2019bidirectional}
C.~Liu, H.~Xie, Z.~Zha, L.~Yu, Z.~Chen, and Y.~Zhang, ``Bidirectional
  attention-recognition model for fine-grained object classification,''
  \emph{IEEE Transactions on Multimedia}, vol.~22, no.~7, pp. 1785--1795, 2019.

\bibitem{rodriguez2018attend}
P.~Rodr{\'\i}guez, J.~M. Gonfaus, G.~Cucurull, F.~XavierRoca, and J.~Gonzalez,
  ``Attend and rectify: a gated attention mechanism for fine-grained
  recovery,'' in \emph{European Conference on Computer Vision (ECCV)}, 2018,
  pp. 349--364.

\bibitem{liu2020filtration}
C.~Liu, H.~Xie, Z.-J. Zha, L.~Ma, L.~Yu, and Y.~Zhang, ``Filtration and
  distillation: Enhancing region attention for fine-grained visual
  categorization,'' in \emph{AAAI Conference on Artificial Intelligence
  (AAAI)}, vol.~34, no.~07, 2020, pp. 11\,555--11\,562.

\bibitem{wu2021align}
D.~Wu, S.~Li, Z.~Zang, K.~Wang, L.~Shang, B.~Sun, H.~Li, and S.~Z. Li, ``Align
  yourself: Self-supervised pre-training for fine-grained recognition via
  saliency alignment,'' \emph{arXiv preprint arXiv:2106.15788}, 2021.

\bibitem{wang2022bridge}
J.~Wang, Y.~Li, X.-S. Wei, H.~Li, Z.~Miao, and R.~Zhang, ``Bridge the gap
  between supervised and unsupervised learning for fine-grained
  classification,'' \emph{arXiv preprint arXiv:2203.00441}, 2022.

\bibitem{chang2021your}
D.~Chang, K.~Pang, Y.~Zheng, Z.~Ma, Y.-Z. Song, and J.~Guo, ``Your" flamingo"
  is my" bird": Fine-grained, or not,'' in \emph{IEEE Conference on Computer
  Vision and Pattern Recognition (CVPR)}, 2021, pp. 11\,476--11\,485.

\bibitem{zhou2020look}
M.~Zhou, Y.~Bai, W.~Zhang, T.~Zhao, and T.~Mei, ``Look-into-object:
  Self-supervised structure modeling for object recognition,'' in \emph{IEEE
  Conference on Computer Vision and Pattern Recognition (CVPR)}, 2020, pp.
  11\,774--11\,783.

\bibitem{kalayeh2018human}
M.~M. Kalayeh, E.~Basaran, M.~G{\"o}kmen, M.~E. Kamasak, and M.~Shah, ``Human
  semantic parsing for person re-identification,'' in \emph{Proceedings of the
  IEEE conference on computer vision and pattern recognition}, 2018, pp.
  1062--1071.

\bibitem{meng2020parsing}
D.~Meng, L.~Li, X.~Liu, Y.~Li, S.~Yang, Z.-J. Zha, X.~Gao, S.~Wang, and
  Q.~Huang, ``Parsing-based view-aware embedding network for vehicle
  re-identification,'' in \emph{Proceedings of the IEEE/CVF conference on
  computer vision and pattern recognition}, 2020, pp. 7103--7112.

\bibitem{zhao2021heterogeneous}
J.~Zhao, Y.~Zhao, J.~Li, K.~Yan, and Y.~Tian, ``Heterogeneous relational
  complement for vehicle re-identification,'' in \emph{Proceedings of the
  IEEE/CVF International Conference on Computer Vision}, 2021, pp. 205--214.

\bibitem{chen2022pedestrian}
W.-C. Chen, X.-Y. Yu, and L.-L. Ou, ``Pedestrian attribute recognition in video
  surveillance scenarios based on view-attribute attention localization,''
  \emph{Machine Intelligence Research}, vol.~19, no.~2, pp. 153--168, 2022.

\bibitem{lake2015human}
B.~M. Lake, R.~Salakhutdinov, and J.~B. Tenenbaum, ``Human-level concept
  learning through probabilistic program induction,'' \emph{Science}, vol. 350,
  no. 6266, pp. 1332--1338, 2015.

\bibitem{fei2006one}
L.~Fei-Fei, R.~Fergus, and P.~Perona, ``One-shot learning of object
  categories,'' \emph{IEEE Transactions on Pattern Analysis and Machine
  Intelligence}, vol.~28, no.~4, pp. 594--611, 2006.

\bibitem{miller2000learning}
E.~G. Miller, N.~E. Matsakis, and P.~A. Viola, ``Learning from one example
  through shared densities on transforms,'' in \emph{IEEE Conference on
  Computer Vision and Pattern Recognition (CVPR)}, vol.~1.\hskip 1em plus 0.5em
  minus 0.4em\relax IEEE, 2000, pp. 464--471.

\bibitem{tang2020revisiting}
L.~Tang, D.~Wertheimer, and B.~Hariharan, ``Revisiting pose-normalization for
  fine-grained few-shot recognition,'' in \emph{Proceedings of the IEEE/CVF
  Conference on Computer Vision and Pattern Recognition}, 2020, pp.
  14\,352--14\,361.

\bibitem{koniusz2021power}
P.~Koniusz and H.~Zhang, ``Power normalizations in fine-grained image, few-shot
  image and graph classification,'' \emph{IEEE Transactions on Pattern Analysis
  and Machine Intelligence}, vol.~44, no.~2, pp. 591--609, 2021.

\bibitem{huang2020low}
H.~Huang, J.~Zhang, J.~Zhang, J.~Xu, and Q.~Wu, ``Low-rank pairwise alignment
  bilinear network for few-shot fine-grained image classification,'' \emph{IEEE
  Transactions on Multimedia}, vol.~23, pp. 1666--1680, 2020.

\bibitem{zhu2020multi}
Y.~Zhu, C.~Liu, and S.~Jiang, ``Multi-attention meta learning for few-shot
  fine-grained image recognition.'' in \emph{IJCAI}, 2020, pp. 1090--1096.

\bibitem{li2019zero}
A.-X. Li, K.-X. Zhang, and L.-W. Wang, ``Zero-shot fine-grained classification
  by deep feature learning with semantics,'' \emph{International Journal of
  Automation and Computing}, vol.~16, no.~5, pp. 563--574, 2019.

\bibitem{Zhang_2016_CVPR}
X.~Zhang, F.~Zhou, Y.~Lin, and S.~Zhang, ``Embedding label structures for
  fine-grained feature representation,'' in \emph{Proceedings of the IEEE
  Conference on Computer Vision and Pattern Recognition}, 2016, pp. 1114--1123.

\bibitem{xie2015hyper}
S.~Xie, T.~Yang, X.~Wang, and Y.~Lin, ``Hyper-class augmented and regularized
  deep learning for fine-grained image classification,'' in \emph{IEEE
  Conference on Computer Vision and Pattern Recognition (CVPR)}, 2015, pp.
  2645--2654.

\bibitem{radford2021learning}
A.~Radford, J.~W. Kim, C.~Hallacy, A.~Ramesh, G.~Goh, S.~Agarwal, G.~Sastry,
  A.~Askell, P.~Mishkin, J.~Clark \emph{et~al.}, ``Learning transferable visual
  models from natural language supervision,'' in \emph{International Conference
  on Machine Learning (ICML)}.\hskip 1em plus 0.5em minus 0.4em\relax PMLR,
  2021, pp. 8748--8763.

\bibitem{li2022grounded}
L.~H. Li, P.~Zhang, H.~Zhang, J.~Yang, C.~Li, Y.~Zhong, L.~Wang, L.~Yuan,
  L.~Zhang, J.-N. Hwang \emph{et~al.}, ``Grounded language-image
  pre-training,'' in \emph{IEEE Conference on Computer Vision and Pattern
  Recognition (CVPR)}, 2022, pp. 10\,965--10\,975.

\bibitem{kalogerakis20173d}
E.~Kalogerakis, M.~Averkiou, S.~Maji, and S.~Chaudhuri, ``3d shape segmentation
  with projective convolutional networks,'' in \emph{IEEE Conference on
  Computer Vision and Pattern Recognition (CVPR)}, 2017, pp. 3779--3788.

\bibitem{yu2019partnet}
F.~Yu, K.~Liu, Y.~Zhang, C.~Zhu, and K.~Xu, ``Partnet: A recursive part
  decomposition network for fine-grained and hierarchical shape segmentation,''
  in \emph{IEEE Conference on Computer Vision and Pattern Recognition (CVPR)},
  2019, pp. 9491--9500.

\bibitem{mo2019partnet}
K.~Mo, S.~Zhu, A.~X. Chang, L.~Yi, S.~Tripathi, L.~J. Guibas, and H.~Su,
  ``Partnet: A large-scale benchmark for fine-grained and hierarchical
  part-level 3d object understanding,'' in \emph{IEEE Conference on Computer
  Vision and Pattern Recognition (CVPR)}, 2019, pp. 909--918.

\bibitem{qi2017pointnet}
C.~R. Qi, H.~Su, K.~Mo, and L.~J. Guibas, ``Pointnet: Deep learning on point
  sets for 3d classification and segmentation,'' in \emph{IEEE Conference on
  Computer Vision and Pattern Recognition (CVPR)}, 2017, pp. 652--660.

\bibitem{kittler20053d}
J.~Kittler, A.~Hilton, M.~Hamouz, and J.~Illingworth, ``3d assisted face
  recognition: A survey of 3d imaging, modelling and recognition approachest,''
  in \emph{2005 IEEE Computer Society Conference on Computer Vision and Pattern
  Recognition (CVPR'05)-Workshops}.\hskip 1em plus 0.5em minus 0.4em\relax
  IEEE, 2005, pp. 114--114.

\bibitem{joung2021learning}
S.~Joung, S.~Kim, M.~Kim, I.-J. Kim, and K.~Sohn, ``Learning canonical 3d
  object representation for fine-grained recognition,'' in \emph{IEEE
  International Conference on Computer Vision (ICCV)}, 2021, pp. 1035--1045.

\bibitem{mildenhall2021nerf}
B.~Mildenhall, P.~P. Srinivasan, M.~Tancik, J.~T. Barron, R.~Ramamoorthi, and
  R.~Ng, ``Nerf: Representing scenes as neural radiance fields for view
  synthesis,'' \emph{Communications of the ACM}, vol.~65, no.~1, pp. 99--106,
  2021.

\end{thebibliography}

\begin{IEEEbiography}[{\includegraphics[width=1in,height=1.25in,clip,keepaspectratio]{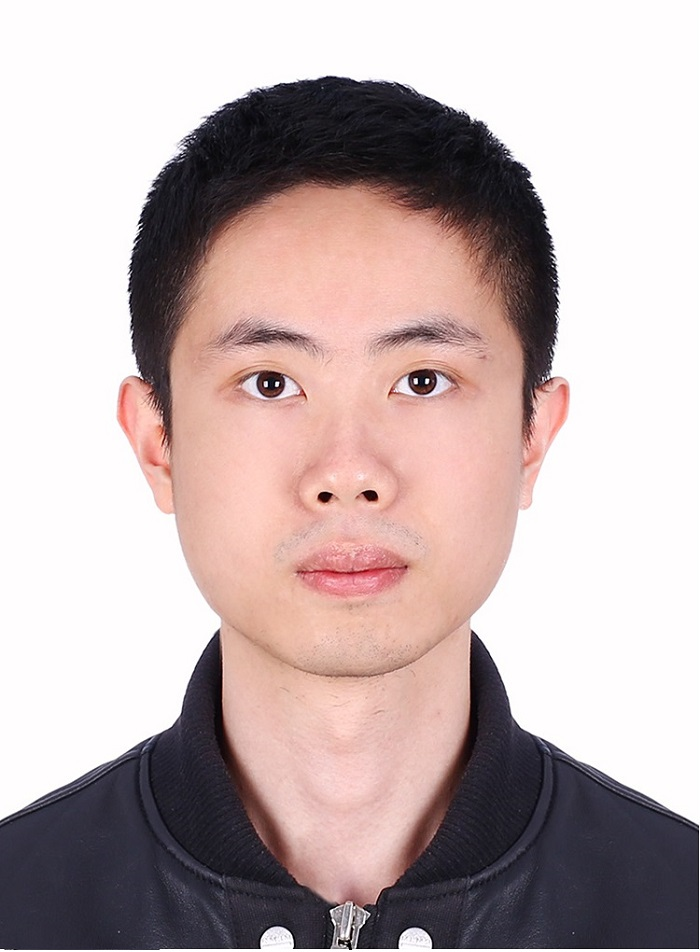}}]{Yifan Zhao} is currently a postdoctoral researcher with the School of Computer Science, Peking University, Beijing, China. He received the B.E. degree from Harbin Institute of Technology in Jul. 2016 and the Ph.D. degree from the School of Computer Science and Engineering, Beihang University, in Nov. 2021. His research interests include computer vision and image/video understanding.
\end{IEEEbiography}

\begin{IEEEbiography}[{\includegraphics[width=1in,height=1.25in,clip,keepaspectratio]{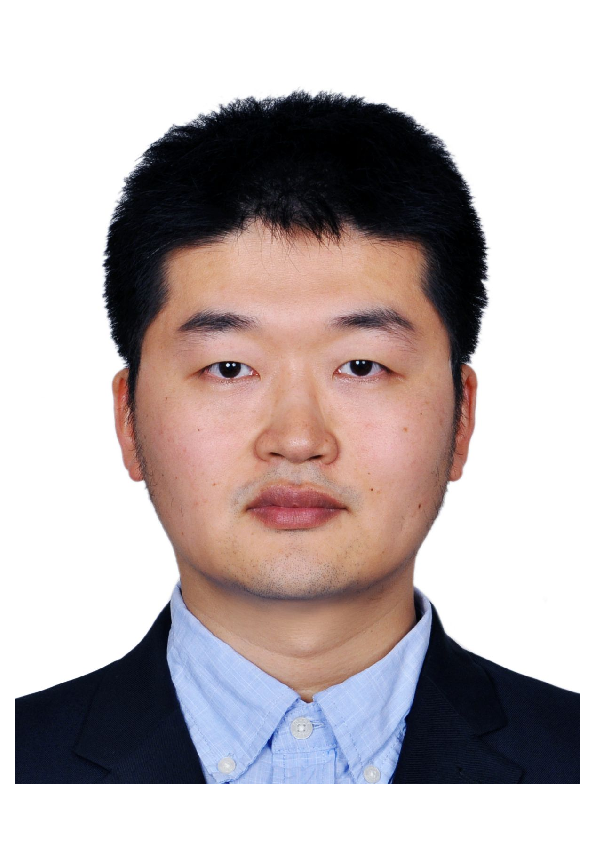}}]{Jia Li} (M'12-SM'15) is currently a Full Professor with the State Key Laboratory of Virtual Reality Technology and Systems, School of Computer Science and Engineering, Beihang University. He received his B.E. degree from Tsinghua University in 2005 and Ph.D. degree from Institute of Computing Technology, Chinese Academy of Sciences, in 2011. Before he joined Beihang University in 2014, he used to work at Nanyang Technological University, Shanda Innovations, and Peking University. His research is focused on computer vision, multimedia and artificial intelligence, especially the visual computing in extreme environments. He has co-authored more than 110 articles in peer-reviewed top-tier journals and conferences. He also has one Monograph published by Springer and more than 60 patents issued from U.S. and China. He is a Fellow of IET, and senior members of IEEE/ACM/CCF/CIE.
\end{IEEEbiography}

\begin{IEEEbiography}[{\includegraphics[width=1in,height=1.25in,clip,keepaspectratio]{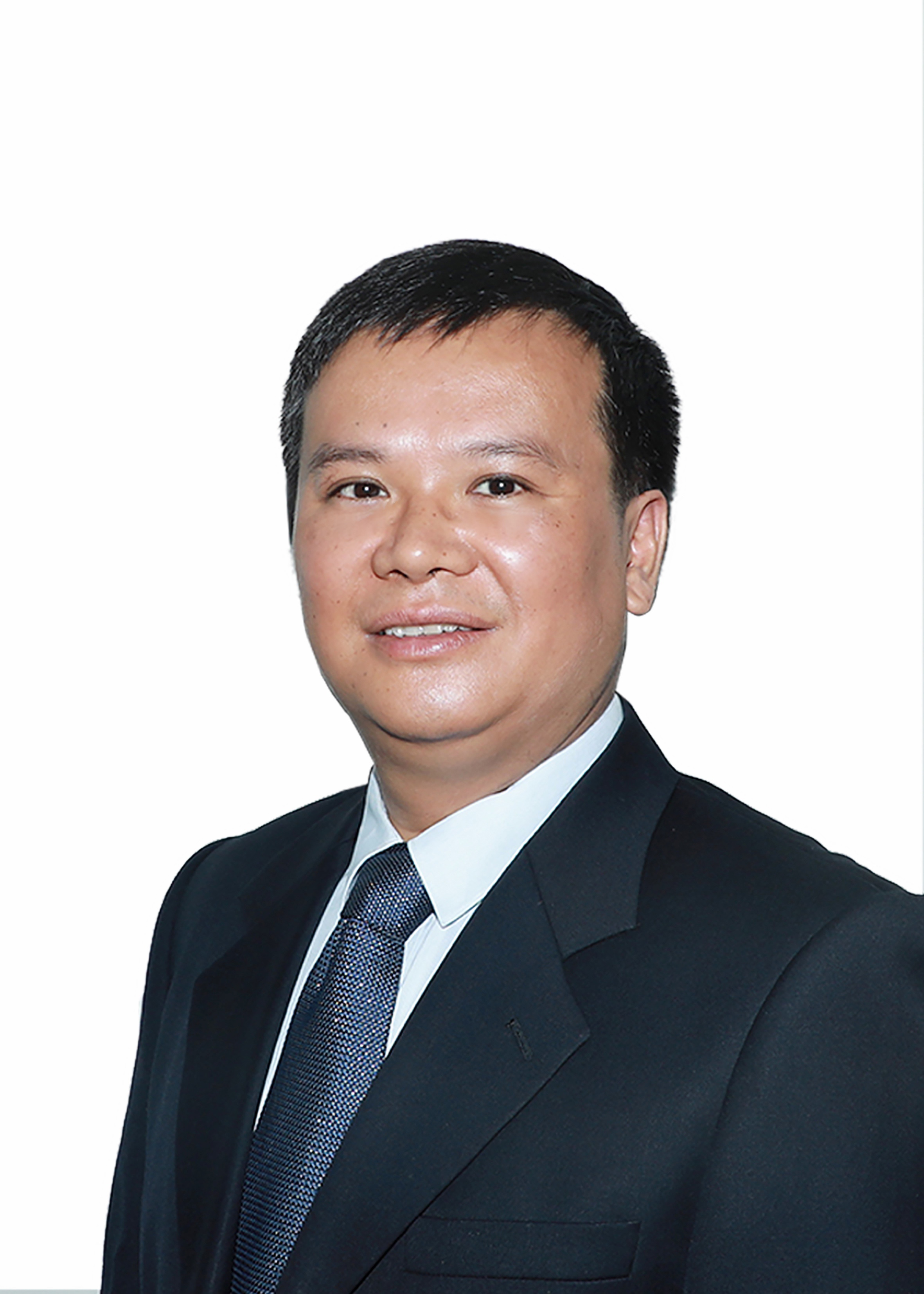}}]{Yonghong Tian} (S'00-M'06-SM'10) is currently a Boya Distinguished Professor with the School of Computer Science, Peking University, China, and is also the deputy director of Artificial Intelligence Research Center, PengCheng Laboratory, Shenzhen, China. His research interests include neuromorphic vision, distributed machine learning and multimedia big data. He is the author or coauthor of over 280 technical articles in refereed journals and conferences. Prof. Tian was/is an Associate Editor of IEEE TCSVT (2018.1-2021.12), IEEE TMM (2014.8-2018.8), IEEE Multimedia Mag. (2018.1-), and IEEE Access (2017.1-). He co-initiated IEEE Intl Conf. on Multimedia Big Data (BigMM) and served as the TPC Co-chair of BigMM 2015, and aslo served as the Technical Program Co-chair of IEEE ICME 2015, IEEE ISM 2015 and IEEE MIPR 2018/2019, and General Co-chair of IEEE MIPR 2020 and ICME 2021. He is the steering member of IEEE ICME (2018-2020) and IEEE BigMM (2015-), and is a TPC Member of more than ten conferences such as CVPR, ICCV, ACM KDD, AAAI, ACM MM and ECCV. He was the recipient of the Chinese National Science Foundation for Distinguished Young Scholars in 2018, two National Science and Technology Awards and three ministerial-level awards in China, and obtained the 2015 EURASIP Best Paper Award for Journal on Image and Video Processing, and the best paper award of IEEE BigMM 2018. He is a Fellow of IEEE, a senior member of CIE and CCF, a member of ACM.
\end{IEEEbiography}

\balance

\end{document}